\documentclass{article}

\PassOptionsToPackage{table}{xcolor}
\usepackage[preprint]{neurips_2026}
\usepackage{microtype}
\usepackage{graphicx}
\usepackage{subcaption}
\usepackage{booktabs} 
\usepackage{xcolor}
\usepackage{colortbl}
\usepackage{multirow}
\usepackage{wrapfig}

\usepackage{amsmath}
\usepackage{amssymb}
\usepackage{mathtools}
\usepackage{amsthm}
\usepackage[normalem]{ulem}

\theoremstyle{plain}

\theoremstyle{definition}

\theoremstyle{remark}

\definecolor{lightblue}{RGB}{120,170,220}
\definecolor{lightred}{RGB}{220,150,150}

\definecolor{latentblue}{RGB}{232,243,255}
\definecolor{discgray}{gray}{0.93}
\usepackage{listings}
\usepackage{tcolorbox}
\tcbuselibrary{listings, breakable}
\usepackage{enumitem}

\newtcolorbox{methodbox}[1]{
    breakable,
    colback=blue!5!white,
    colframe=blue!75!black,
    title=#1,
    fonttitle=\bfseries
}

\newtcolorbox{examplebox}[1]{
    breakable,
    colback=green!5!white,
    colframe=green!60!black,
    title=#1,
    fonttitle=\bfseries
}

\newtcolorbox{vaelatexbox}[1]{
    breakable,
    colback=purple!5!white,
    colframe=purple!75!black,
    title=#1,
    fonttitle=\bfseries
}

\usepackage[utf8]{inputenc}
\usepackage[T1]{fontenc}
\usepackage{hyperref}
\usepackage{url}
\usepackage{amsfonts}
\usepackage{nicefrac}
\usepackage{array}
\usepackage{siunitx}

\definecolor{codeblue}{HTML}{2E86C1}
\definecolor{codered}{HTML}{C0392B}
\definecolor{codegreen}{HTML}{27AE60}

\newcommand{\cmark}{\textcolor{codegreen}{\checkmark}}
\newcommand{\xmark}{\textcolor{codered}{$\times$}}
\definecolor{rowgray}{HTML}{F7F7F7}

\title{LaDi-RL: Latent Diffusion Reasoning Prevents Entropy Collapse in Reinforcement Learning}

\author{%
  Haoqiang Kang\\
  UC San Diego\\
  \And
  Yizhe Zhang\\ 
  Apple\\
  \And
  Nikki Lijing Kuang\\
  UC San Diego\\
  \And
  Yi-An Ma\\
  UC San Diego\\
  \And
  Lianhui Qin\\
  UC San Diego\\
}

\begin{document}

\maketitle

\begin{abstract}

Reinforcement learning has become a central paradigm for improving LLM reasoning, but most existing methods optimize policies over discrete token sequences. This creates a mismatch between the optimization space and the structure of reasoning: many important decisions are semantic, global, and trajectory-level rather than local token choices. Continuous latent-space RL offers a promising alternative by allowing policies to explore higher-level reasoning representations. However, simply moving to latent space is not sufficient. The resulting policy must model a complex, multi-modal distribution over valid reasoning trajectories. We therefore propose \textbf{La}tent \textbf{Di}ffusion Reasoning with \textbf{R}einforcement \textbf{L}earning \textbf{(LaDi-RL)}, where a diffusion model generates latent reasoning trajectories through iterative denoising. This formulation enables structured exploration and expressive distribution modeling, but also introduces a fundamental credit-assignment challenge: the policy acts in latent space, while rewards are observed only after the latent is decoded into text. A naive rollout strategy therefore entangles latent reasoning quality with text decoding quality, making it unclear whether an incorrect answer results from a poor latent trajectory or from an imperfect textual realization. To address this, we introduce hierarchical latent-text rollouts. We sample multiple text completions for each latent trajectory and aggregate their rewards to obtain a decoder-marginalized estimate of latent utility. This provides a cleaner and lower-variance reward signal for optimizing the diffusion policy. Empirically, LaDi-RL outperforms token-level RL by 9.4\% on code generation and 5.7\% on math reasoning in pass@1, and even surpasses the base model’s pass@k performance. 

\end{abstract}

\vspace{-5mm}
\section{Introduction}
\vspace{-2mm}

\begin{wrapfigure}{r}{0.6\linewidth}
    \centering
    \vspace{-5mm}
    \includegraphics[width=\linewidth]{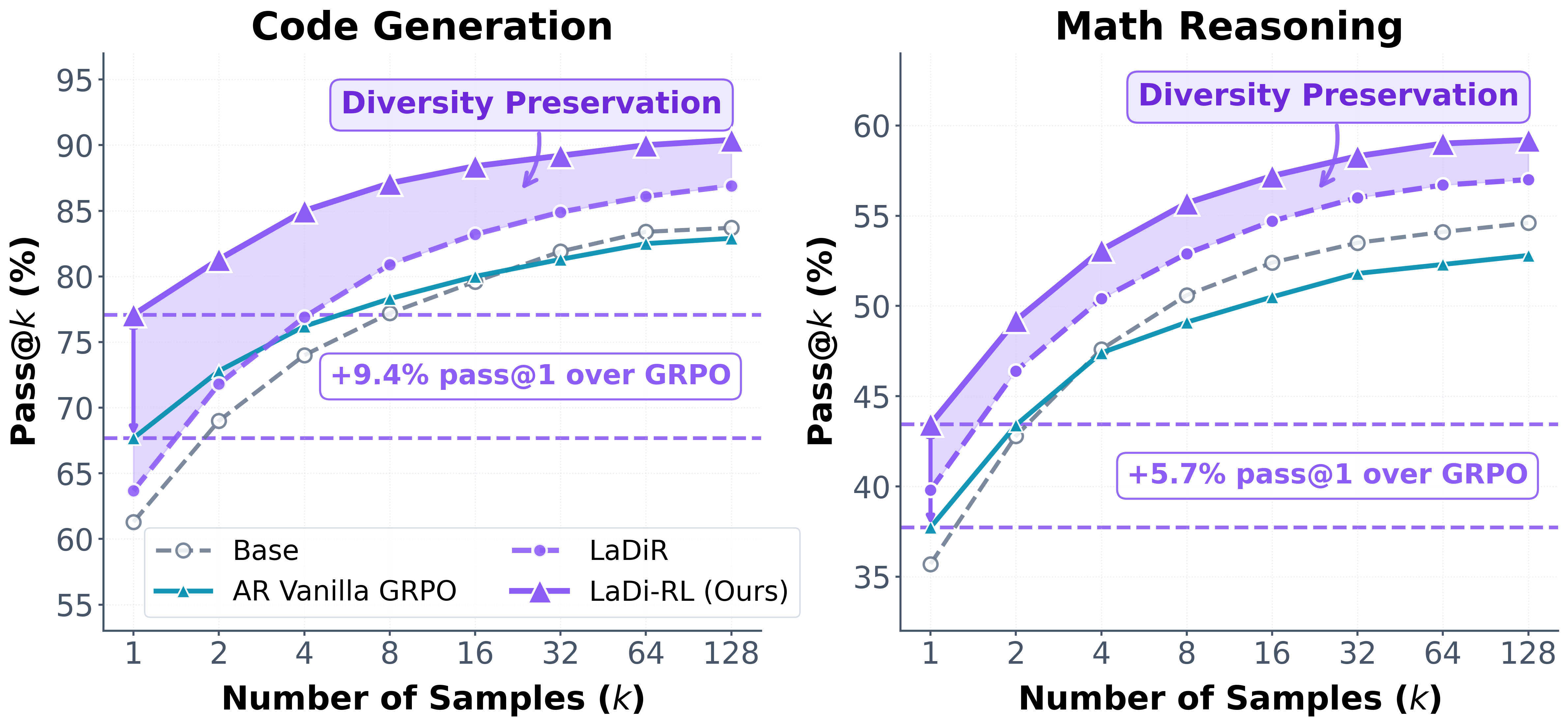}
    \vspace{-6mm}
    \caption{\textbf{Average pass@$k$ over all tested benchmarks.} LaDi-RL preserves solution diversity and converts LaDiR's pass@$k$ gains into strong pass@$1$ improvements, whereas vanilla GRPO exhibits diversity collapse at large $k$.}
    \vspace{-3.5mm}
    \label{fig:teaser}
\end{wrapfigure}
Reinforcement learning (RL) has become a dominant paradigm for improving the reasoning ability of large language models (LLMs), particularly by optimizing the generation of reasoning trajectories~\citep{guo2025deepseek, yu2025dapo, shao2024deepseekmath}. However, this process can suffer from \textit{diversity collapse}~\citep{song2024mind, dang2025weight, yue2025does, zhao2025echo, he2025rewarding}: during training, the model increasingly concentrates probability mass on a narrow set of reasoning patterns while suppressing alternative ones~\citep{song2024mind, dang2025weight, yue2025does, zhao2025echo, he2025rewarding}. This collapse presents a critical bottleneck for further RL optimization: when the model's sample distribution becomes more concentrated, the explored trajectories become less diverse and the resulting RL performance gains become limited. This phenomenon is visible in pass@$k$ metrics~\citep{he2025rewarding, cobbe2021training, yue2025does, chung2025revisiting, chen2025rethinking}: while RL-tuned models often improve pass@$1$, they may underperform the original base model at large $k$, indicating a loss of diverse correct trajectories, as shown by vanilla GRPO in Figure~\ref{fig:teaser}.

Recent work mitigates this problem by encouraging \textit{semantic diversity} during training~\citep{yao2025diversity, hu2025diversity, li2025jointly, naik2023diversity, chen2025pass, walder2025pass}. These methods typically retain token-level exploration and add diversity-oriented rewards or regularizers. We argue that this creates a fundamental mismatch: \textit{semantic diversity is a global, trajectory-level property, but the LLM policy explores through local token-level perturbations.} As a result, the model can easily generate lexical variants of the same reasoning trajectory, such as replacing ``therefore'' with ``so'' or rephrasing an intermediate step, without discovering semantically different strategies such as switching from an algebraic derivation to a geometric argument. Token-level exploration therefore is an indirect and inefficient mechanism for optimizing semantic diversity.

To address this mismatch, we propose \textbf{La}tent \textbf{Di}ffusion Reasoning with \textbf{R}einforcement \textbf{L}earning \textbf{(LaDi-RL)}. LaDi-RL moves exploration from the token level to the trajectory level using latent reasoning, where an entire reasoning trajectory is encoded into continuous latent tokens~\citep{kang2025ladir}. These latent tokens then condition the autoregressive generation of the final answer text, which are evaluated by downstream task rewards. In this latent space, perturbing the latent tokens changes the underlying reasoning trajectory as a whole, allowing the model to explore semantically different solution strategies rather than merely surface-form lexical variations.

Effective latent-space exploration, however, requires an expressive policy. Reasoning trajectories form a complex and multi-modal distribution: a problem may admit many valid solution strategies, each corresponding to a different region of latent space. Simple continuous policies may struggle to represent this structure or may average across modes, producing incoherent trajectories. We therefore parameterize the latent policy as a diffusion model over trajectory representations~\citep{ho2020denoising, song2019generative}. Starting from noise, the policy can iteratively denoise toward a latent reasoning trajectory, injecting stochasticity throughout generation while converging towards the data distribution. This process supports structured exploration and naturally enables diversity-promoting guidance, such as repulsion during denoising~\citep{d2021repulsive, zilberstein2024repulsive}.

Optimizing such a latent policy introduces a new credit-assignment challenge. The policy acts in latent space, but rewards are observed only after the latent trajectory is decoded into text and evaluated on the downstream task. A failed solution may therefore arise from a poor latent reasoning trajectory or from an imperfect textual realization of an otherwise useful latent. Naively assigning the reward of a single decoded completion to the latent trajectory entangles latent quality with decoder noise, yielding a noisy and potentially misleading learning signal.

We address this issue with \textit{hierarchical latent-text rollouts}. Under a fixed rollout budget, instead of sampling only independent latent-text pairs, LaDi-RL first samples a smaller set of latent reasoning trajectories and then decodes multiple answer texts from each latent. The rewards of these text completions are aggregated to estimate the latent trajectory’s decoder-marginalized downstream utility, i.e., how useful the latent is on average across possible textual realizations. This reduces the chance of incorrectly penalizing a good latent because of a single failed decoding, or incorrectly rewarding a weak latent because of a lucky completion. As a result, hierarchical rollouts provide a cleaner, lower-variance learning signal for optimizing the latent diffusion policy.

We evaluate LaDi-RL on code generation and mathematical reasoning benchmarks, where it consistently improves both accuracy and diversity compared to token-level RL baselines. It achieves absolute pass@$1$ improvements of +9.4\% and +5.7\% respectively while simultaneously surpassing the base model's pass@$k$ performance. These results demonstrate that diffusion-based latent trajectory optimization, combined with decoder-marginalized reward estimation, provides a principled alternative to token-level RL for improving LLM reasoning without sacrificing solution diversity.

\vspace{-3mm}
\section{Preliminary}
\vspace{-2mm}
\subsection{Group Relative Policy Optimization}
\vspace{-1.5mm}
Group Relative Policy Optimization (GRPO)~\cite{guo2025deepseek} is a reinforcement learning algorithm that has been widely adopted for optimizing large generative models, including large language models~\cite{yu2025dapo} and diffusion-based generative models~\cite{liu2025flow, xue2025dancegrpo}. Formally, given a group of rollout trajectories $\{\mathbf{o}_g\}_{g=1}^G$ sampled from the current policy $\pi_\theta$, where the trajectories share the same conditioning input (e.g., prompt or context), GRPO computes a normalized advantage for each trajectory directly from scalar rewards:
$\hat{A}_g = \frac{R_g - \mathrm{mean}(\{R_g\}_{g=1}^G)}{\mathrm{std}(\{R_g\}_{g=1}^G)}$
where $R_g$ denotes the outcome reward of trajectory $\mathbf{o}_g$. The GRPO objective is defined as:
\begin{equation}
\resizebox{0.93\hsize}{!}{%
$\mathcal{J}_{\mathrm{GRPO}}(\theta)
={}
\mathbb{E}_{\{\mathbf{o}_g\}_{g=1}^G \sim \pi_{\theta_{\mathrm{old}}}}
\!\Bigg[
\frac{1}{G} \sum_{g=1}^G
\min\!\Big(
r_g(\theta)\hat{A}_g,
\mathrm{clip}\big(r_g(\theta), 1-\epsilon, 1+\epsilon\big)\hat{A}_g
\Big)
\Bigg] 
-
\beta\, D_{\mathrm{KL}}\!\left(\pi_\theta \,\|\, \pi_{\theta_{\mathrm{ref}}}\right).$}
\label{eq:grpo_obj}
\end{equation}
where $ r_g(\theta) = \frac{\pi_\theta(\mathbf{o}_g)}{\pi_{\theta_{\mathrm{old}}}(\mathbf{o}_g)} $
and $\pi_{\theta_{\mathrm{ref}}}$ is a fixed reference policy, $\epsilon$ is the clipping threshold, and $\beta$ controls KL regularization. 



\vspace{-1.5mm}


\vspace{-1.5mm}
\paragraph{GRPO for Flow Matching Models.}
Flow-GRPO~\cite{liu2025flow, li2025mixgrpo} applies GRPO to flow matching models by converting the deterministic Flow-ODE into an equivalent stochastic differential equation (SDE) and discretizing it using the Euler--Maruyama scheme. This introduces controlled stochasticity while preserving the underlying flow structure. Specifically, the resulting transition kernel takes the form:
\vspace{-1.5mm}
\begin{equation}
\pi_\theta(\mathbf{x}_{t-1} \mid \mathbf{x}_t)
=
\mathcal{N}\!\left(
\mathbf{x}_{t-1};
\mu_\theta(\mathbf{x}_t),
g_t^2 \Delta t \, \mathbf{I}
\right),
\label{eq:flow_transition}
\vspace{-1mm}
\end{equation}
where $g_t = a \sqrt{\frac{t}{1-t}}$ controls the noise scale, $\Delta t$ is the discretization step size, and the mean $\mu_\theta(\mathbf{x}_t)$ is determined by the learned flow field $v_\theta(\mathbf{x}_t, t)$ following the standard flow-matching discretization. This formulation reduces each transition to a tractable Gaussian distribution, enabling direct computation of likelihood ratios and seamless integration with the GRPO objective in Eq.~\eqref{eq:grpo_obj}. Further details are provided in Appendix~\ref{sec:prelim_appendix}.
\vspace{-1mm}
\subsection{Latent Diffusion Reasoner (LaDiR)}
\vspace{-1.5mm}
LaDiR~\cite{kang2025ladir} is a latent diffusion–based reasoning framework that compresses text CoTs into a compact continuous latent block, enabling efficient semantic-level reasoning. In this work, we adopt LaDiR as a \emph{cold-start} initialization for subsequent reinforcement learning.
\begin{figure*}[h]
    \centering
    \vspace{-2mm}
    \includegraphics[width=1\linewidth]{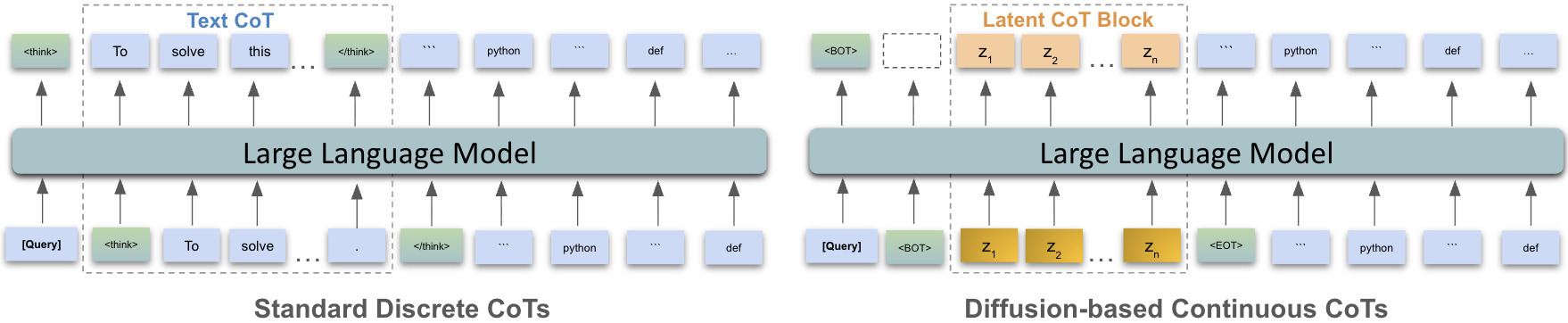}
    \vspace{-3mm}
    \caption{Comparison between standard discrete CoT reasoning and latent continuous CoT reasoning. Left: conventional autoregressive generation produces reasoning directly in token space. Right: reasoning is represented as a fixed-size latent block, enabling continuous modeling and decoupling reasoning from surface text generation.}
    \label{fig:dis_cont_comparison}
    \vspace{-5mm}
\end{figure*}

\vspace{-2mm}
\paragraph{Architecture.} As illustrated in Figure~\ref{fig:dis_cont_comparison}, LaDiR represents a reasoning process using a single continuous latent CoT block. Given a question $Q$, we insert a special token \texttt{<BOT>} to indicate the beginning of the latent block, followed by a fixed number of latent tokens $Z = \{z_1, \dots, z_B\}$, and a token \texttt{<EOT>} to mark its end. The pretrained LLM then generates the final answer autoregressively conditioned on both the question and the latent block. The latent tokens are produced by compressing a short reference CoT (typically fewer than 1k tokens) using a variational autoencoder (VAE), yielding a fixed-size latent representation (e.g., $B=64$). Unlike the original LaDiR formulation, which applies blockwise diffusion over multiple latent segments, we employ a single sufficiently large latent block, which we find adequate for reconstructing short CoTs while simplifying the model.
\vspace{-2mm}
\paragraph{Training Objective.}
LaDiR is trained using a joint objective that aligns latent reasoning with downstream text generation. Let $Z^{\mathrm{enc}}$ denote the latent tokens encoded from a reference CoT. We train a conditional latent generator using a flow-matching objective: $\mathcal{L}_{\mathrm{FM}}
= \mathbb{E} \left[
\| v_\theta(Z_t,Q,t)-(Z^{\mathrm{enc}}-\epsilon)\|_2^2
\right]$
, where $Z_t=(1-t)\epsilon+tZ^{\mathrm{enc}}$. Conditioned on the question $Q$ and latent block $Z$, the LLM is simultaneously trained to generate the target answer $Y$ using a standard autoregressive cross-entropy loss:
$\mathcal{L}_{\mathrm{CE}} = - \mathbb{E}_{(Q,Z,Y)} \sum_t
\log p_\theta(y_t \mid y_{<t}, Q, Z). $ The final training objective is $\mathcal{L}
= \lambda \mathcal{L}_{\mathrm{FM}} + \mathcal{L}_{\mathrm{CE}} $, where $\lambda$ is loss weight for the latent part.
\vspace{-3mm}
\paragraph{Inference.}
During inference, latent tokens are generated by iterative denoising conditioned on the input question and then the model generates the answer text given the denoised latent block.
\vspace{-3mm}
\section{Methodology}
\vspace{-1.5mm}






Building upon LaDiR, we propose \emph{Latent Diffusion Reasoner with Reinforcement Learning (LaDi-RL)}, a framework that applies GRPO to latent diffusion policies, enabling structured and flexible exploration in continuous latent space with higher diversity. A complementary text policy then conditions on the denoised latent reasoning trajectories to generate the final answer text.


\begin{figure*}[t]
    \vspace{-3mm}
    \centering
    \includegraphics[width=\linewidth]{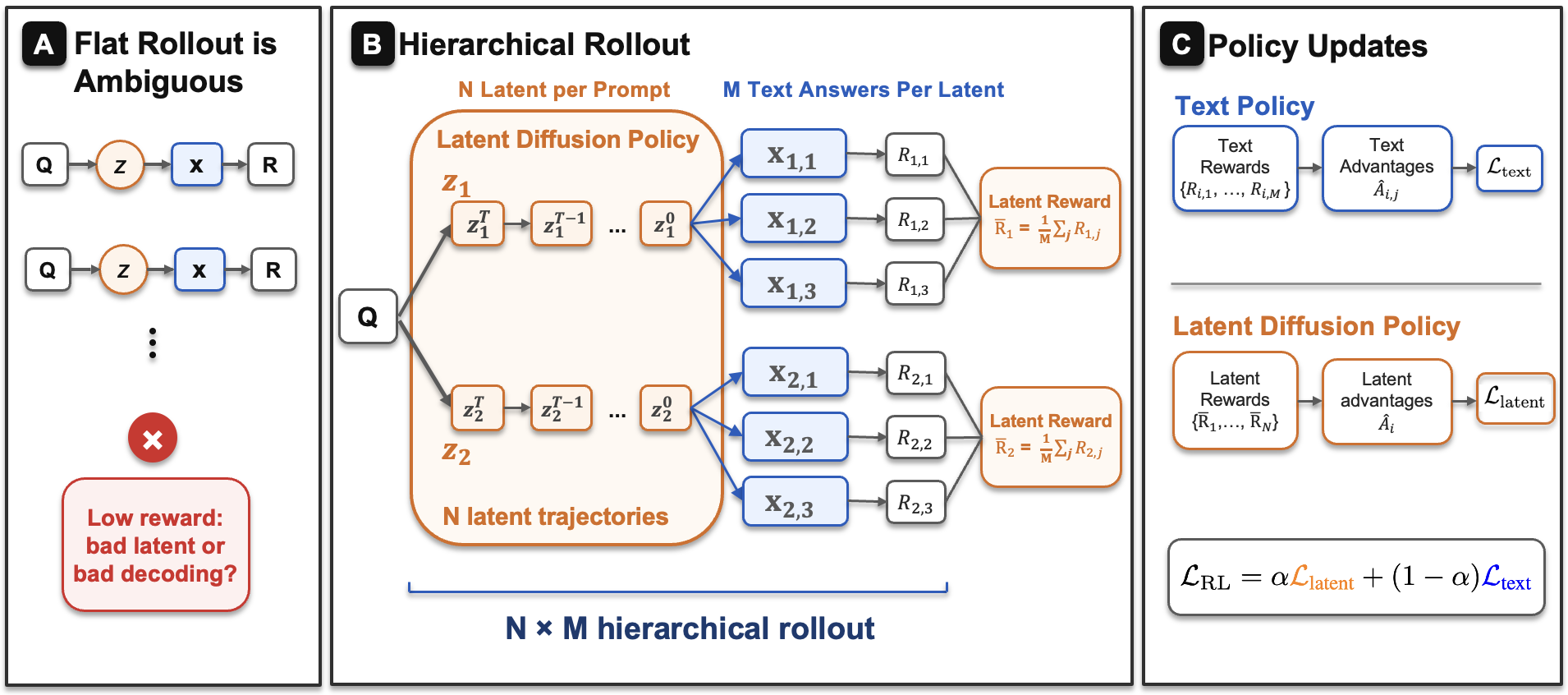}
    \vspace{-4mm}
    \caption{\textbf{Overview of LaDi-RL.}
    \textbf{(a)} Problem of standard flat rollout.
    \textbf{(b)} \emph{Hierarchical $N{\times}M$ rollout:} the latent diffusion policy samples $N$ latent blocks $\{z_n^0\}$, and the text policy generates $M$ candidate answers $\{x_{n,m}\}$ per latent.
    \textbf{(c)} \emph{Policy updates:} the latent policy is trained on the \emph{average} reward of the $M$ texts under each latent, and the text policy on \emph{relative} differences among those $M$ texts; both are optimized with GRPO.
    }
    \label{fig:pipeline}
    \vspace{-5mm}
\end{figure*}

\vspace{-1mm}
\subsection{Hierarchical Rollout}
\vspace{-2mm}
LaDi-RL involves two heterogeneous policies operating at different levels of abstraction: a latent diffusion policy first samples a continuous latent reasoning trajectory, and a text decoder then realizes that latent trajectory into an answer. This factorization enables trajectory-level exploration, but it also introduces a sharper credit-assignment challenge than standard token-level LLM reasoning. The downstream reward is observed only after the latent trajectory has been decoded into text, so the reward of a single latent-text rollout conflates two distinct factors: whether the latent trajectory encodes a useful reasoning mode, and whether the text decoder faithfully realizes that mode into a correct answer. As shown in Figure~\ref{fig:pipeline}~(a), this makes credit assignment ambiguous: a low reward cannot tell whether the latent reasoning is flawed or whether a useful latent was merely decoded into a poor answer, making it difficult to optimize the latent and text policies separately.

To address this, we propose a \emph{hierarchical $N{\times}M$ rollout} (Figure~\ref{fig:pipeline}~(b)). Given a query $Q$, a single trajectory is represented as $\mathbf{o} = \{\mathbf{z}^1, \ldots, \mathbf{z}^K, x^{K+1}, \ldots, x^{K+L}\}$, where $\mathbf{z}^i$ is the $i$-th latent block generated by the latent diffusion model via $K$ iterative denoising steps (forming a latent CoT), and $x^{K+1}, \ldots, x^{K+L}$ are $L$ autoregressively generated text tokens conditioned on the final latent block. For each query we first sample $N$ latent diffusion trajectories $\{\mathbf{z}_{n}\}_{n=1}^N$, and then sample $M$ text responses $\{\mathbf{x}_{(n,m)}\}_{m=1}^M$ conditioned on each latent. Rewards are computed at the final answer text. This hierarchy disentangles the two sources of stochasticity: averaging the $M$ rewards under each latent isolates the \emph{quality of the latent reasoning}, while their relative differences isolate \emph{which generated text works best for that reasoning}—giving the latent and text policies clean, complementary learning signals from the same set of rollouts.

For efficiency, we use a small number of denoising steps $K$ (e.g., $K{=}10$) and disable classifier-free guidance~\cite{ho2022classifier} during rollout.
\vspace{-3mm}
\subsection{Latent Diffusion Policy}
\vspace{-2mm}
We optimize the latent diffusion policy following FlowGRPO \cite{liu2025flow}, turning deterministic ODE denoising step to SDE step as a stochastic policy step. The latent policy loss follows the GRPO formulation:
\vspace{-2mm}
\begin{align}
\mathcal{L}_{\mathrm{latent}}^{\mathrm{clip}(\epsilon_z)} = \frac{1}{N}& \sum_{n=1}^{N} \sum_{t=1}^{K} \min \Big( r_{n,t}(\phi) \hat{A}_n,
\operatorname{clip}(r_{n,t}(\phi), 1-\epsilon_z, 1+\epsilon_z) \hat{A}_n \Big),
\label{eq:grpo}
\vspace{-7mm}
\end{align} where $r_{n,t}(\phi) = \frac{p_\phi\left(\mathbf{z}_n^{t} \mid \mathbf{z}_{n}^{t-1}, Q\right)}{p_{\phi_{\mathrm{old}}}\left(\mathbf{z}_n^{t} \mid \mathbf{z}_{n}^{t-1}, Q\right)}$ and the corresponding group-relative advantage $\hat{A}_{n}$ is computed by standardizing the $N$ mean rewards $\{\bar{R}_{n}\}_{n=1}^N$ across the latent blocks for the same query $Q$, where each $\bar{R}_{n}$ is calculated by averaging over the $M$ text answers $\mathbf{x}_{(n,m)}$ conditioned on the same $n$-th latent block: $ \bar{R}^{(n)} = \frac{1}{M} \sum_{m=1}^{M} R(\mathbf{x}_{(n,m)})$, where $R$ is the outcome reward function on answer text.

\vspace{-2mm}
\begin{wrapfigure}{r}{0.43\linewidth}
    \vspace{-5mm}
    \centering
    \includegraphics[width=\linewidth]{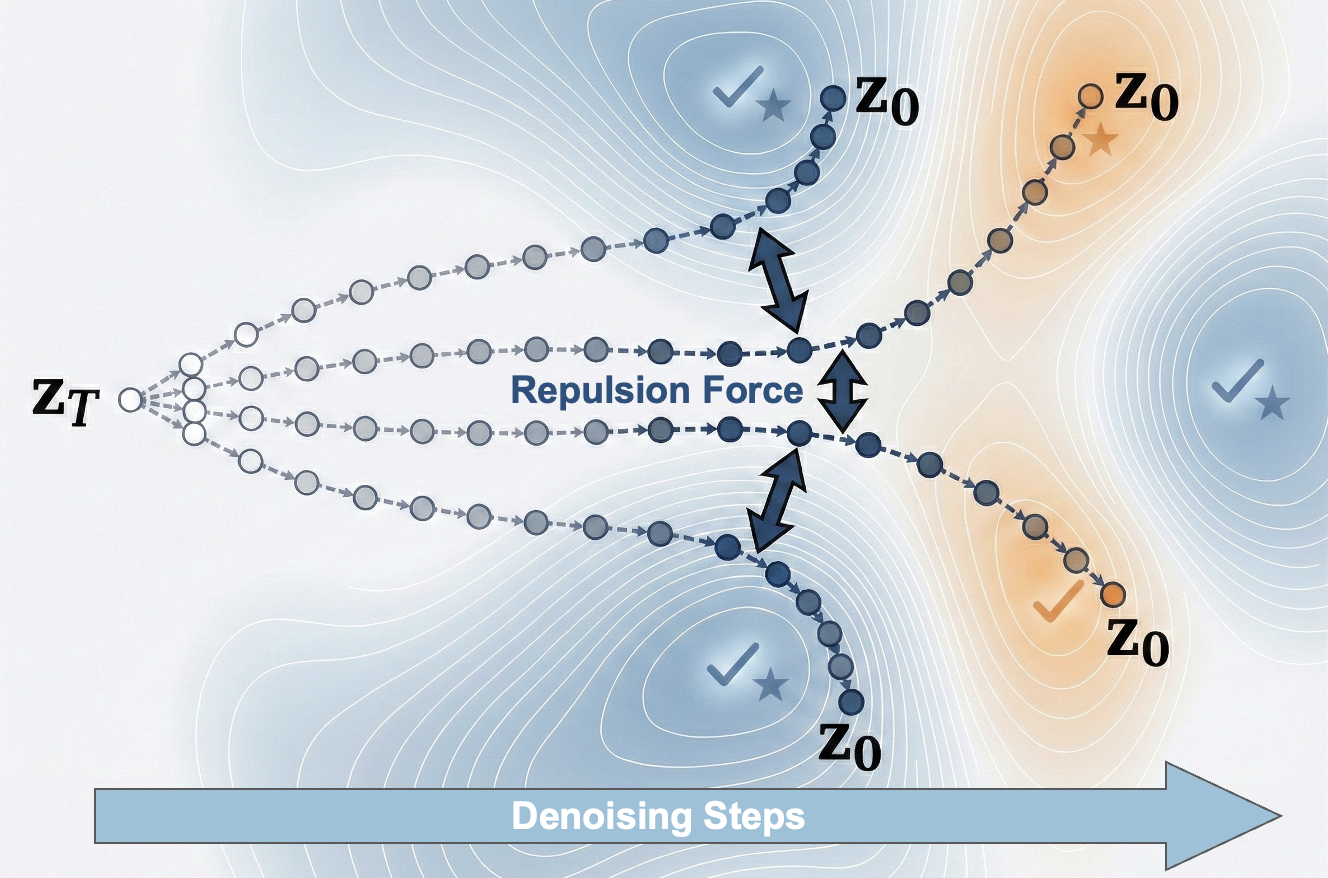}
    \caption{\textbf{Latent diffusion-based exploration with diversity guidance.} Multi-step denoising maps a noise to diverse solutions, while a repulsive force pushes trajectories apart latent space to encourage exploration.}
    \vspace{-6mm}
    \label{fig:diversity_guide}
\end{wrapfigure}
  
\vspace{-1mm}
\paragraph{Diversity Guidance}
To further improve diversity during the rollout process, inspired by kernel-density estimation and repulsive forces~\cite{d2021repulsive, zilberstein2024repulsive}, we introduce a repulsion-based guidance mechanism (see Figure~\ref{fig:diversity_guide}). At each denoising step $t$, given the group of $N$ latent trajectories $\{\mathbf{z}_{n}^t\}_{n=1}^N$ sampled for a query $Q$, we first compute a bandwidth parameter:
$ \sigma = \mathrm{median}_{n < n'} \|\mathbf{z}_{n}^t - \mathbf{z}_{n'}^t\|_2. $
The repulsion force on a latent block \noindent $\mathbf{z}_{n}^t$ is defined by the interaction force $\mathbf{F}(\mathbf{z}_{n}^t)$:
\vspace{-1mm}
\begin{equation}
    \mathbf{F}(\mathbf{z}_{n}^t)= \sum_{n' \neq n} 2 \left( 1 - \frac{d_{nn'}^2}{2\sigma^2} \right) \exp \left( -\frac{d_{nn'}^2}{2\sigma^2} \right) (\mathbf{z}_{n}^t - \mathbf{z}_{n'}^t)
\end{equation}
where $d_{nn'}^2 = \|\mathbf{z}_{n}^t - \mathbf{z}_{n'}^t\|_2^2$. This encourages separation among nearby latent trajectories while suppressing interactions between distant ones. The repulsion is applied with higher magnitude at early denoising steps and progressively reduced as $t \to 0$. Specifically, we define a time-dependent scale $\gamma_t = \gamma_{\max} \frac{t}{K}$, where $K$ is the total number of denoising steps. The diversity-guided latent update is then:
\vspace{-1mm}
\begin{equation}
\hat{\mathbf{z}}_{n}^{t-1} = f_\phi(\mathbf{z}_{n}^t, t, Q) + \gamma_t \mathbf{F}(\mathbf{z}_{n}^t),
\vspace{-0mm}
\end{equation}
where $f_\phi(\mathbf{z}_{n}^t, t, Q)$ denotes the base diffusion model prediction. This repulsion-based guidance serves as a form of local geometric regularization, explicitly separating nearby latent trajectories, while the latent diffusion process provides global multi-modal support. Together, they reshape the geometry of exploration, preserving diversity during RL training without relying on policy entropy.
\vspace{-0mm}

\vspace{-2mm}
\subsection{Text Policy and Joint Optimization.}
\vspace{-2mm}
The text policy $p_\theta$ acts as a complementary module, grounding the reasoning generated by the latent diffusion policy into specific text answers. We employ a local standard GRPO loss for it:
\begin{equation}
 \mathcal{L}_{\mathrm{text}}^{\mathrm{clip}(\epsilon_x^{l}, \epsilon_x^{h})} 
 = \frac{1}{NM} \sum_{n=1}^{N} \sum_{m=1}^{M} \sum_{j=1}^{L} \min \Big( r_{n,m,j}(\theta) \hat{A}_{n,m} ,
 \operatorname{clip}(r_{n,m,j}(\theta), 1-\epsilon_x^{l}, 1+\epsilon_x^{h})\hat{A}_{n,m}  \Big),
\end{equation}
where $r_{n,m,j}(\theta) = \frac{p_\theta(x_{n,m,j} \mid x_{n,m,<j}, \mathbf{z}_n)}{p_{\theta_{\mathrm{old}}}(x_{n,m,j} \mid x_{n,m,<j}, \mathbf{z}_n)}$ is the importance sampling ratio for the $j$-th token, and $\hat{A}_{n,m} = \frac{R(y_{n, m}) - \text{mean}(\{R(y_{n, i})\}_{i=1}^M)}{\text{std}(\{R(y_{n, i})\}_{i=1}^M)}$ represents the local group advantage calculated strictly within the $M$ text samples generated conditioned on the same latent block $\mathbf{z}_n$ . This design decouples text generation from the broader latent-space exploration, preventing the text policy from being penalized for poor latent trajectories and instead forcing it to find the optimal text sequences for the reasoning provided. Then to joint train the two policies, we have this in the final objective:
\vspace{-1mm}
\begin{equation}
\mathcal{L}_{\mathrm{RL}} = \alpha \mathcal{L}_{\mathrm{latent}}^{\mathrm{clip}(\epsilon_z)} + (1-\alpha) \mathcal{L}_{\mathrm{text}}^{\mathrm{clip}(\epsilon_x^{l}, \epsilon_x^{h})},
\vspace{-0mm}
\end{equation}

$\mathcal{L}_{\mathrm{text}}$
$\mathcal{L}_{\mathrm{latent}}$
where $\alpha$ is a balancing coefficient that prioritizes the optimization of the primary latent reasoning space while maintaining the text-alignment task. As illustrated in the training pipeline (Figure \ref{fig:pipeline}), this joint loss ensures that as the latent diffusion policy $p_\phi$ explores diverse reasoning chains-of-thought, the text policy $p_\theta$ simultaneously adapts to serve as a faithful and effective text generation.

\vspace{-3mm}
\section{Experiment}
\vspace{-2mm}
We evaluate our method on two representative reasoning domains: code generation and math reasoning. We compare against strong baselines under controlled settings and report standard pass@$1$ and pass@$k$ metrics across benchmarks. See more experimental details in Appendix~\ref{app:exp_details}.

\vspace{-3mm}
\subsection{Experimental Setup}
\vspace{-2mm}

\paragraph{Dataset} For code generation, we initialize the model with SFT on the Ling-Coder dataset~\cite{codefuse2025samplemattersleveragingmixtureofexperts}, which contains 1.4M Python-only samples. RL training is then performed on 24k filtered problems drawn from AceCoder~\cite{AceCoder} and KodCoder~\cite{xu2025kodcode}. We evaluate model performance on HumanEval~\cite{chen2021evaluating}, MBPP~\cite{austin2021program}, their extended variants HumanEval+ and MBPP+~\cite{liu2023your}, as well as LiveCodeBench v6 (LCB V6)~\cite{jain2024livecodebench}. For math reasoning, we use the R1-distill dataset~\cite{slam-distillation-from-r1} for SFT and the DeepScaleR-Preview-Dataset~\cite{deepscaler2025} for RL training, which contains approximately 40K unique problem–answer pairs. Evaluation is conducted on six challenging benchmarks: AIME 2024~\cite{aime_1983_2024}, AIME 2025~\cite{aime25}, AMC 2023, MATH-500~\cite{hendrycks2021measuring}, Minerva Math~\cite{lewkowycz2022solving}, and OlympiadBench~\cite{he2024olympiadbench}.
\vspace{-3mm}
\paragraph{Baselines} We compare our method against a diverse set of strong baselines spanning autoregressive, diffusion-based, and reinforcement learning approaches. For code generation, we include open-sourced AR models such as Qwen2.5-Coder~\cite{hui2024qwen2}, OpenCoder~\cite{huang2025opencoder}, rStar-Coder~\cite{liu2025rstar}, OlympicCoder~\cite{openr1}, and Seed-Coder~\cite{seed2025seed}, alongside reasoning-enhanced models like OpenThinker~\cite{guha2025openthoughts} and DeepSeek-R1-Distill~\cite{guo2025deepseek}. We also compare with discrete diffusion language models including Dream~\cite{ye2025dream}, Dream-Coder~\cite{xie2025dream}, LLaDA~\cite{nie2025large}, and Diffu-Coder~\cite{gong2025diffucoder}. We include looped latent reasoning model Ouro~\cite{zhu2025scaling} to highlight the benefits of diffusion models. To validate our training methods, we compare against various reasoning methods—including Soft Thinking~\cite{zhang2025soft}, Standard SFT, and TaH+~\cite{fu2025think}—and reinforcement learning baselines such as AR vanilla GRPO~\cite{guo2025deepseek}, GRPO with entropy advantages~\cite{cheng2025reasoning}, HybridGRPO~\cite{sane2025hybrid}, SofT-GRPO~\cite{zheng2025soft} and Multiplex Thinking~\cite{tang2026multiplex}. In addition, for math reasoning, we focus on representative methods including vanilla GRPO~\cite{guo2025deepseek}, GRPO with entropy advantages~\cite{cheng2025reasoning}, Soft Thinking~\cite{zhang2025soft}, and Multiplex Thinking~\cite{tang2026multiplex} to enable controlled comparisons with prior approaches. We also include two baselines by replacing flow matching loss with standard $\ell_2$ loss on VAE (LaVAE) and apply GRPO on it (LaVAE-RL).

\vspace{-3mm}
\paragraph{Implementation Details}
For code generation, we use \textit{Qwen3-Base}~\cite{yang2025qwen3} as the base model, while for math reasoning we adopt \textit{DeepSeek-R1-Distill-Qwen-7B}~\cite{guo2025deepseek}. To mitigate performance variance inherent in single-run evaluations, we sample 16 solutions per problem and report the average pass@$1$ accuracy for all benchmarks. See Appendix~\ref{app:exp_details} for more details.
\vspace{-4mm}

\section{Results}

\vspace{-3mm}
In this section, we present the main results, analysis, and ablation studies on the two benchmarks. See efficiency analysis, more ablation study, and qualitative results in Appendix~\ref{app:add_results}.

\begin{table*}[ht!]
\vspace{-2mm}
\centering
\small
\setlength{\tabcolsep}{6pt}

\newcommand{\pos}[1]{\textcolor{green!60!black}{$_{+#1}$}}
\newcommand{\negnew}[1]{\textcolor{red!70!black}{$_{-#1}$}}

\resizebox{\textwidth}{!}{%
\begin{tabular}{lccccccc}
\toprule
\textbf{Method}
& \textbf{AIME24}
& \textbf{AIME25} 
& \textbf{AMC23} 
& \textbf{MATH500} 
& \textbf{Minerva} 
& \textbf{Olympiad} 
& \textbf{Average} \\
\midrule

\multicolumn{8}{c}{\cellcolor{gray!10}\textbf{Non-RL Methods}} \\

\textcolor{gray!60}{Base (DS-R1-Distill-Qwen-7B) } 
& \textcolor{gray!60}{15.7}
& \textcolor{gray!60}{16.0}
& \textcolor{gray!60}{42.4}
& \textcolor{gray!60}{71.6}
& \textcolor{gray!60}{33.3}
& \textcolor{gray!60}{35.6}
& \textcolor{gray!60}{35.77} \\

Soft Thinking
& 20.3\pos{4.6}
& 19.1\pos{3.1}
& 47.9\pos{5.5}
& 76.5\pos{4.9}
& 37.2\pos{3.9}
& 40.6\pos{5.0}
& 40.27\pos{4.50} \\

LaVAE
& 6.5\negnew{9.2}
& 7.8\negnew{8.2}
& 21.0\negnew{21.4}
& 47.5\negnew{24.1}
& 17.6\negnew{15.7}
& 18.2\negnew{17.4}
& 19.77\negnew{16.0} \\

LaDiR
& 18.8\pos{3.1}
& 19.8\pos{3.8}
& 49.2\pos{6.8}
& 78.4\pos{6.8}
& 38.9\pos{5.6}
& 41.0\pos{5.4}
& 41.01\pos{5.24} \\

\midrule
\multicolumn{8}{c}{\cellcolor{gray!10}\textbf{RL Training Methods}} \\

AR Vanilla GRPO
& 17.2\pos{1.5}
& 17.1\pos{1.1}
& 44.7\pos{2.3}
& 74.1\pos{2.5}
& 35.3\pos{2.0}
& 38.0\pos{2.4}
& 37.73\pos{1.96} \\

GRPO+Entropy Adv.
& 18.0\pos{2.3}
& 18.8\pos{2.8}
& 43.1\pos{0.7}
& 76.8\pos{5.2}
& 37.8\pos{4.5}
& 38.4\pos{2.8}
& 38.82\pos{3.05} \\

Multiplex Thinking
& 20.6\pos{4.9}
& 19.7\pos{3.7}
& 50.7\pos{8.3}
& 78.0\pos{6.4}
& 38.6\pos{5.3}
& 41.7\pos{6.1}
& 41.55\pos{5.78} \\

LaVAE-RL
& 7.3\negnew{8.4}
& 8.5\negnew{7.5}
& 26.6\negnew{15.8}
& 53.4\negnew{18.2}
& 20.9\negnew{12.4}
& 26.8\negnew{8.8}
& 23.92\negnew{11.85} \\

\textbf{LaDi-RL}
& \textbf{22.3}\pos{6.6}
& \textbf{20.5}\pos{4.5}
& \textbf{52.6}\pos{10.2}
& \textbf{81.4}\pos{9.8}
& \textbf{40.7}\pos{7.4}
& \textbf{43.2}\pos{7.6}
& \textbf{43.45}\pos{7.68} \\

\bottomrule
\end{tabular}%
}

\caption{
Results on math reasoning benchmarks.
We report pass@$1$ accuracy, with relative improvements over the base model shown as subscript in \textcolor{green!60!black}{green} and negative changes in \textcolor{red!70!black}{red}.
}
\label{tab:math_reasoning_results}
\vspace{-1mm}
\end{table*}

\vspace{-3mm}
\begin{table*}[ht!]
\centering
\small
\setlength{\tabcolsep}{6pt}

\definecolor{groupgray}{gray}{0.93}
\definecolor{methodbeige}{RGB}{245,240,230}
\definecolor{discgray}{gray}{0.97}   
\definecolor{latentblue}{RGB}{232,243,255} 
\definecolor{methodblue}{RGB}
{205,220,240}   
\definecolor{methodyellow}{RGB}{238,220,180}

\definecolor{sectiongray}{gray}{0.93}        
\newcommand{\pos}[1]{\textcolor{green!60!black}{$_{+#1}$}}
\newcommand{\negnew}[1]{\textcolor{red!70!black}{$_{-#1}$}}

\newcommand{\discrow}{\rowcolor{discgray}}
\newcommand{\latentrow}{\rowcolor{latentblue}}

\resizebox{\textwidth}{!}{%
\begin{tabular}{@{}l@{\hspace{2pt}}c l c c c c c}
\toprule
\textbf{Model / Method}
& \textbf{Base Model}
& \textbf{MBPP}
& \textbf{MBPP+}
& \textbf{HumanEval}
& \textbf{HumanEval+}
& \textbf{LCB V6}
& \textbf{Avg.} \\
\midrule

\multicolumn{8}{c}{\cellcolor{methodyellow}\textbf{Open-Sourced Models}} \\
\multicolumn{8}{c}{\cellcolor{groupgray}Autoregressive Coding Models} \\

Qwen 2.5 Coder$^*$ 
& Qwen2.5-7B
& 75.9 & 62.9 & 66.5 & 60.4 & 26.8 & 59.90 \\

OpenCoder$^*$ 
& from scratch (8B)
& 79.9 & 70.4 & 66.5 & 63.4 & 29.6 & 61.96 \\

rStar-Coder$^*$ 
& Qwen2.5-7B 
& 87.9 & 74.0 & 95.9 & 90.8 & 53.5 & 80.42 \\

OlympicCoder$^*$ 
& Qwen2.5-7B 
& 80.0 & 66.4 & 82.1 & 76.9 & 37.3 & 68.54 \\

DeepSeek-R1-Distill$^*$ 
& Qwen2.5-7B 
& 78.4 & 66.7 & 89.6 & 83.7 & 34.2 & 70.52 \\

OpenThinker2$^*$ 
& Qwen2.5-7B 
& 86.9 & 73.9 & 92.7 & 87.8 & 29.2 & 74.10 \\

Seed-Coder$^*$ 
& from scratch 
& 82.0 & 69.0 & 77.4 & 68.3 & 28.4 & 65.02 \\
\midrule
\multicolumn{8}{c}{\cellcolor{groupgray}Diffusion Language Models} \\

Dream$^*$ 
& Qwen2.5-7B 
& 68.7 & 57.4 & 56.7 & 50.0 & 18.6 & -- \\

LLaDA$^*$ 
& from scratch (8B)
& 50.1 & 42.1 & 35.4 & 30.5 & 12.4 & -- \\

Diffu-Coder$^*$ 
& Qwen2.5-Coder-7B 
& 75.1 & 61.9 & 72.0 & 65.2 & 24.5 & 59.74 \\

Dream-Coder$^*$ 
& Qwen2.5-Coder-7B
& 75.9 & 61.6 & 66.5 & 60.4 & 21.4 & 57.16 \\

d1$^*$
& LLaDA+RL 
& 39.0
& --
& 45.5
& --
& --
& -- \\

\midrule
\multicolumn{8}{c}{\cellcolor{groupgray}Looped Latent Reasoning Models} \\

Ouro$^*$ 
& from scratch (2.6B)
& 80.4 & 66.6 & 78.2 & 70.7 & 38.7 & 66.92 \\

\midrule


\multicolumn{8}{c}{\cellcolor{methodyellow}\textbf{Method Comparison}} \\
\multicolumn{8}{c}{\cellcolor{sectiongray}Non-RL Methods}\\

\textcolor{gray!60}{Base Model}
& Qwen3-8B-Base
& \textcolor{gray!60}{60.5}
& \textcolor{gray!60}{53.8}
& \textcolor{gray!60}{78.2}
& \textcolor{gray!60}{68.6}
& \textcolor{gray!60}{37.8}
& \textcolor{gray!60}{60.58} \\

Standard SFT
& Qwen3-8B-Base
& 63.3\pos{2.8}
& 52.7\negnew{1.1}
& 84.6\pos{6.4}
& 69.5\pos{0.9}
& 39.5\pos{1.7}
& 61.32\pos{0.74} \\

\latentrow
Soft Thinking
& Qwen3-8B-Base
& 64.2\pos{3.7}
& 53.1\negnew{0.7}
& 85.0\pos{6.8}
& 71.2\pos{2.6}
& --
& -- \\

\latentrow
TaH+
& Qwen3-8B-Base
& 65.6\pos{5.1}
& 56.5\pos{2.7}
& 85.8\pos{7.6}
& 74.3\pos{5.7}
& --
& -- \\

\latentrow
LaVAE
& Qwen3-8B-Base
& 42.0\negnew{18.5}
& 30.2\negnew{23.6}
& 47.8\negnew{30.4}
& 32.8\negnew{35.8}
& 10.8\negnew{27.0}
& 32.72\negnew{27.86} \\

\latentrow
LaDiR
& Qwen3-8B-Base
& 66.8\pos{6.3}
& 59.5\pos{5.7}
& 87.4\pos{9.2}
& 73.2\pos{4.6}
& 41.0\pos{3.2}
& 65.58\pos{5.00} \\

\midrule
\multicolumn{8}{c}{\cellcolor{gray!10}RL Training Methods} \\

AR Vanilla GRPO
& Qwen3-8B-Base
& 72.3\pos{11.8}
& 61.7\pos{7.9}
& 82.6\pos{4.4}
& 74.2\pos{5.6}
& 47.6\pos{9.8}
& 67.68\pos{7.10} \\

GRPO w/ Entropy Adv.
& Qwen3-8B-Base
& 73.8\pos{13.3}
& 62.9\pos{9.1}
& 84.1\pos{5.9}
& 77.8\pos{9.2}
& 48.2\pos{10.4}
& 69.36\pos{8.78} \\

\latentrow
HybridGRPO
& Qwen3-8B-Base
& 73.1\pos{12.6}
& 62.0\pos{8.2}
& 82.9\pos{4.7}
& 76.8\pos{8.2}
& 48.0\pos{10.2}
& 68.56\pos{7.98} \\

\latentrow
Soft Token
& Qwen3-8B-Base
& 70.2\pos{9.7}
& 60.6\pos{6.8}
& 81.7\pos{3.5}
& 70.4\pos{1.8}
& 40.9\pos{3.1}
& 64.76\pos{4.18} \\

\latentrow
SofT-GRPO
& Qwen3-8B-Base
& 72.8\pos{12.3}
& 60.2\pos{6.4}
& 83.9\pos{5.7}
& 72.2\pos{3.6}
& 41.6\pos{3.8}
& 66.14\pos{5.56} \\

\latentrow
LaVAE-RL
& Qwen3-8B-Base
& 45.6\negnew{14.9}
& 30.9\negnew{22.9}
& 53.2\negnew{25.0}
& 34.0\negnew{34.6}
& 12.5\negnew{25.3}
& 35.24\negnew{25.34} \\

\latentrow
\textbf{LaDi-RL}
& Qwen3-8B-Base
& \textbf{84.2}\pos{23.7}
& \textbf{75.1}\pos{21.3}
& \textbf{90.5}\pos{12.3}
& \textbf{82.9}\pos{14.3}
& \textbf{52.7}\pos{14.9}
& \textbf{77.08}\pos{16.50} \\

\bottomrule
\end{tabular}%
}

\vspace{2pt}
\caption{
pass@$1$ results on code generation benchmarks. The subscripts indicate relative change from the \textcolor{gray!60}{Base Model} (\textcolor{green!60!black}{green} for gains, \textcolor{red!70!black}{red} for drops). \colorbox{latentblue}{Blue-shaded} rows indicate latent reasoning methods; unshaded rows denote discrete token-space methods.
$^*$Results are from the original papers.
}
\label{tab:coding_results}
\vspace{-2mm}
\end{table*}

\subsection{Main Results}
\vspace{-1.5mm}
\paragraph{pass@$1$ improvements over baselines.}
Tables~1 and~\ref{tab:math_reasoning_results} show that \textbf{LaDi-RL} consistently outperforms prior \emph{latent reasoning} and \emph{reinforcement learning} methods across both code generation and math reasoning tasks. Compared to latent reasoning approaches such as Soft Thinking, LaDi-RL achieves average improvements of $+3.2\%$ on math reasoning and up to $+4.9\%$ on individual datasets (e.g., MATH500). Comparing to RL baselines, LaDi-RL improves over the latent RL method Multiplex Thinking, by $+1.9\%$ on math reasoning and GRPO with entropy advantage by $+7.8\%$ on code generation, while yielding even larger margins over vanilla GRPO ($+5.7\%$ on math and $+9.4\%$ on code). These results demonstrate that explicitly optimizing over latent diffusion trajectories provides stronger gains than other latent RL or discrete RL alone.

We further compare LaDi-RL with existing well-trained open-sourced coding models. LaDi-RL outperforms all compared models except \emph{rStar-Coder}, matching its performance within a $3.3\%$ absolute margin on average while requiring over $20\times$ fewer training examples. Compared to \emph{diffusion language models}, LaDi-RL shows a large performance gap, improving average pass@$1$ by approximately $+17\%$, with gains around $+18\%$ on HumanEval and $+28\%$ on LiveCodeBench-V6. 
Against \emph{looped latent reasoning models}, LaDi-RL improves average pass@$1$ by $+10.9\%$, with consistent gains across all benchmarks. More broadly, these results suggest that LaDi-RL offers a general framework for improving LLM reasoning performance.
\vspace{-4mm}
\begin{figure*}[htbp!]
    \centering
    \vspace{-1mm}
    \includegraphics[width=\linewidth]{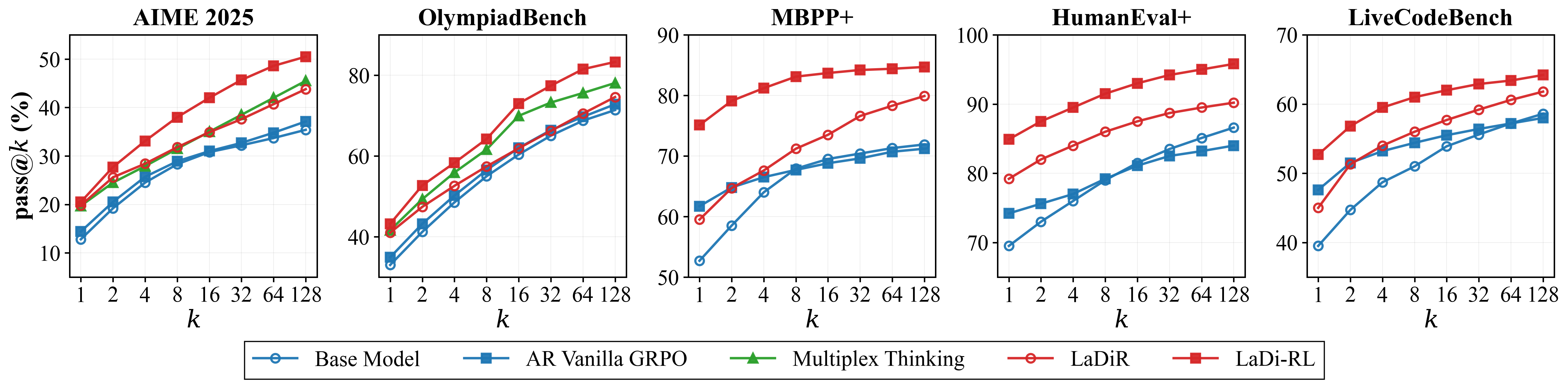}
    \vspace{-5.5mm}
    \caption{pass@$k$ performance on code generation and math reasoning across different $k$.}
    \label{fig:passk}
    \vspace{-7mm}
\end{figure*}
\vspace{-0mm}
\paragraph{Latent diffusion exploration mitigates diversity collapse in pass@$k$.} 
As shown in Figure~\ref{fig:passk}, our results confirm the diversity collapse of AR vanilla GRPO~\cite{yue2025does} in discrete token space: while GRPO improves pass@$1$, it fails to surpass the base model at large $k$, indicating that more samples does not yield broader solution coverage. While recent continuous latent reasoning methods such as \textit{Multiplex Thinking} partially alleviate this issue, our LaDi-RL breaks the base model upper bound at large $k$ on all benchmarks by a much larger margin, achieving absolute improvements of +13.4 (AIME 2025), +11.9 (OlympiadBench), +12.8 (MBPP+), +11.8 (HumanEval+), and +5.6 (LiveCodeBench) at k=128. These results show that latent diffusion exploration mitigates the diversity collapse issue through a multi-step denoising process with guidance and improves the reasoning boundary of base model.

\begin{wraptable}{r}{0.47\linewidth}
    \centering
    \small
    \setlength{\tabcolsep}{3pt}
    \vspace{-10mm}
    \begin{tabular}{lccc}
    \toprule
    \textbf{Method} & \textbf{p@1} & \textbf{p@128} & \textbf{Sem.} \\
    \midrule
    Base Model             & 60.6 & 76.4 & 0.29 \\
    \midrule
    \rowcolor{discgray}\multicolumn{4}{l}{\textit{Token-level exploration}} \\
    Vanilla GRPO (discrete)              & 67.7 & 73.2 & 0.18 \\
    SofT-GRPO (cont. emb.)       & 66.1 & 71.4 & 0.22 \\
    \midrule
    \rowcolor{latentblue}\multicolumn{4}{l}{\textit{Trajectory-level exploration}} \\
    LaVAE-RL (VAE)        & 35.2 & 44.6 & 0.40 \\
    \textbf{LaDi-RL (Diffusion)} & \textbf{77.1} & \textbf{86.1} & \textbf{0.52} \\
    \bottomrule
    \end{tabular}
    \caption{\textbf{Exploration-space ablation} on code benchmarks (avg).}
    \vspace{-3mm}
    \label{tab:ablation_exploration}
    \end{wraptable}

\vspace{-3mm}
\subsection{Ablation Studies}
\vspace{-2mm}
\paragraph{Token-level vs.\ Trajectory-level Exploration.} We compare RL exploration in three spaces: discrete tokens (GRPO~\cite{guo2025deepseek}), continuous token embeddings (SofT-GRPO~\cite{zheng2025soft}), and our latent trajectory space. We report \emph{semantic diversity} (Sem.) as the average percentage of sampled reasoning trajectories judged by GPT-5 to represent distinct solution strategies. As shown in Table~\ref{tab:ablation_exploration}, both token-level variants produce low semantic diversity ($0.18$--$0.22$) and their pass@$k$ saturates well below the base model. Lifting exploration into latent trajectory space finds more semantically different trajectories (Sem.\ up to $0.40$ and $0.52$) and unlocks pass@$1$ gains of $+9.4$ together with $+12.9$ on pass@$128$, confirming our claim that trajectory-level exploration discovers semantically distinct reasoning trajectories with higher semantic scores than token-level exploration, which translates into stronger pass@$k$.
\vspace{-2mm}
\begin{wraptable}{r}{0.96\linewidth}
    \centering
    \small
    \setlength{\tabcolsep}{2pt}
    \vspace{-6mm}
    \begin{tabular}{lcccc}
    \toprule
    \textbf{Method}
    & \textbf{MBPP}
    & \textbf{HE}
    & \textbf{LCB}
    & \textbf{Avg} \\
    \midrule
    LaDiR                       & 66.8 & 87.4 & 41.0 & 65.1 \\
    Text Only (tied)            & 68.2 & 85.9 & 43.1 & 65.7 \\
    Text Only (uncoupled)       & 70.3 & 86.8 & 44.8 & 67.3 \\
    Latent Only (tied)          & 54.3 & 62.6 & 21.6 & 46.2 \\
    Latent Only (uncoupled)     & 73.8 & 89.8 & 47.2 & 70.3 \\
    \midrule
    Latent + Text               & \textbf{84.2} & \textbf{92.2} & \textbf{52.7} & \textbf{76.4} \\
    - w/o div.\ guidance         & 80.1 & 90.3 & 49.4 & 73.3 \\
    \bottomrule
    \end{tabular}
    \caption{Ablation (\textbf{pass@$1$}) on policy composition and diversity guidance. HE: HumanEval; LCB: LCB V6.}
    \vspace{-6mm}
    \label{tab:ablation_latent_text_pass1}
\end{wraptable}
\vspace{-3.5mm}
\paragraph{Diffusion vs.\ VAE Policy in Latent Space.} Within the latent space (blue rows in Table~\ref{tab:ablation_exploration}), we compare LaDi-RL with LaVAE-RL~\cite{butt2025soft, qiu2025noisygrpo}, which uses a single-step VAE sampler over the same latent tokens. Diffusion learns latent reasoning more effectively, yielding both higher semantic diversity (Sem.\ $0.52$ vs.\ $0.40$) and substantially stronger performance ($+41.9$ pass@$1$ and $+41.5$ pass@$128$).
\vspace{-3mm}
\paragraph{Latent exploration is the primary driver; text policy is complementary.} We decompose LaDi-RL's policy into latent-only, text-only, and joint variants in Table~\ref{tab:ablation_latent_text_pass1}. Two findings stand out. First, latent-only optimization (uncoupled, $70.3$ avg) substantially outperforms text-only optimization (uncoupled, $67.3$ avg), confirming that trajectory-level exploration carries the bulk of the gain in LaDi-RL. We note that a latent-only policy with tied weights collapses because the shared backbone receives insufficient supervision from latent rollouts alone; uncoupling the latent policy from the answer-generating copy resolves this. Second, jointly training both policies ($76.4$ avg) further improves over the latent-only setting by $+6.1$ avg, indicating that the text policy provides complementary grounding signals for latent reasoning.
\vspace{-3mm}
\paragraph{Diversity guidance.}
As shown in Table~\ref{tab:ablation_latent_text_pass1}, starting from the best \textit{Latent + Text Policies} setting, removing diversity guidance leads to consistent absolute performance drops of 4.1, 1.9, and 3.3 points. Beyond final performance, Figure~\ref{fig:diversity_analysis} in Appendix further illustrates the impact of diversity guidance on training dynamics: ablating diversity guidance results in noticeably reduced reward variance and slower improvement of rollout reward across both math reasoning and code generation.
\vspace{-3mm}
\paragraph{Hierarchical Rollout.} Table~\ref{tab:ablation_rollout} shows that hierarchical rollouts are more effective for LaDi-RL than for token-level GRPO. While GRPO gains only $+2.2$ pass@1 from the same $4{\times}4$ structure, LaDi-RL improves from $60.0$ to $77.1$ ($+17.1$). This suggests that the hierarchy matches LaDi-RL's factorized generation process, where $N$ explores different latent reasoning modes and $M$ marginalizes over text-level decoding noise for each mode. The ablation further shows that the two axes are not interchangeable. Increasing $N$ alone gives a large gain ($60.0{\to}68.5$), indicating that semantic diversity in latent space is the main driver. Increasing $M$ alone is much weaker ($60.0{\to}63.5$), since repeatedly decoding from a single latent cannot recover missing reasoning modes. However, $M$ becomes valuable when paired with diverse latents: the full $4{\times}4$ rollout reaches $77.1$, suggesting that multi-decode averaging provides a cleaner latent-level reward once there are meaningful latent candidates to compare.

\vspace{-4mm}
\begin{wraptable}{r}{0.48\linewidth}
\centering
\small
\setlength{\tabcolsep}{5pt}
\vspace{-12mm}
\begin{tabular}{cccc}
\toprule
\textbf{Hier.} & \textbf{$N$ (Latent)} & \textbf{$M$ (Text)} & \textbf{pass@1} \\
\midrule
\rowcolor{discgray}\multicolumn{4}{l}{\textit{Token-level Vanilla GRPO}} \\
\xmark & 1 & 1 & 65.5 \\
\cmark & 4 & 4 & 67.7 \\
\midrule
\rowcolor{latentblue}\multicolumn{4}{l}{\textit{LaDi-RL}} \\
\xmark & 1 & 1 & 60.0 \\
\cmark & 4 & 1 & 68.5 \\
\cmark & 1 & 4 & 63.5 \\
\cmark & \textbf{4} & \textbf{4} & \textbf{77.1} \\
\bottomrule
\end{tabular}
\caption{\textbf{Hierarchical-rollout ablation} on code benchmarks (avg pass@$1$). Hier.: \cmark\ if rollout uses $N{\times}M$ hierarchy.}
\vspace{-5mm}
\label{tab:ablation_rollout}
\end{wraptable}
\vspace{-0mm}
\section{Related Works}
\vspace{-3mm}
\paragraph{Latent Reasoning} Latent reasoning methods address the limitations of token-level CoT by shifting reasoning processes into a latent space. Early approaches utilized discrete special tokens to expand internal reasoning capacity or encode implicit intermediate steps, yielding more abstract representations~\citep{herel2024thinkingtokenslanguagemodeling, pfau2024letsthinkdotdot, wang2024guiding, zelikman2024quietstar, zhou2025coevolutionary, jin2025disentanglingmemoryreasoningability}. Subsequent work along this direction demonstrated that reasoning via continuous latent representations rather than discrete tokens can further improve performance. By operating on “soft” tokens that are either self-generated or produced by auxiliary models, LLMs can exploit richer semantic information~\citep{gozeten2025continuous, cheng2024compressedchainthoughtefficient, hao2024traininglargelanguagemodels, liu2024expeditingelevatinglargelanguage, shen2025codicompressingchainofthoughtcontinuous, tack2025llmpretrainingcontinuousconcepts, zhu2025reasoning, butt2025soft, zhang2025soft, wu2025llms}. Furthermore, recent research leverages the expressive power of diffusion models to propose and refine reasoning trajectories \citep{kang2025ladir, anon2024diffusion, shao2025diffuse, venkatraman2024latent, anon2024diffusion, lovelacestop, zhang2023planner, lovelace2023latent, lovelace2024diffusion}. In parallel, a complementary line of work utilizes recurrent or looped architectures to induce latent reasoning internally, bypassing the need to explicitly represent reasoning steps as output tokens~\citep{chen2025innerthinkingtransformerleveraging, geiping2025scalingtesttimecomputelatent, mohtashami2025cotformer, saunshi2025reasoninglatentthoughtspower, yu2025enhancingautoregressivechainofthoughtloopaligned}. 
Despite the above progress, recent efforts apply RL to continuous latent spaces by injecting stochasticity into token embeddings~\cite{butt2025soft, sane2025hybrid, zheng2025soft, tang2026multiplex,ozeren2025reinforcement} or by regulating exploration during policy optimization~\cite{zhang2025relax}. Comparing to these method, our method injects noise via a latent diffusion process, enabling exploration over reasoning trajectories rather than token variations.
\vspace{-4mm}
\paragraph{RL for LLMs} The advancement of Large Reasoning Models has been significantly propelled by RL paradigms, particularly Reinforcement Learning with Verifiable Rewards (RLVR) \citep{guo2025deepseek, openai2024o1, zhang2025survey}. While foundational algorithms like PPO \citep{schulman2017ppo} and DPO \citep{rafailov2023dpo} laid the groundwork for alignment, recent efforts have focused on pure RL methods such as GRPO \citep{guo2025deepseek} and its variants \citep{lin2025cppo, zhang2025stepwise, li2025adaptive} elicit emergent reasoning patterns like self-reflection and verification \citep{deepseek2024v3}. However, a persistent challenge in scaling RLVR is the phenomenon of \textit{diversity collapse}, where the policy rapidly loses diversity and converges to suboptimal local minima \citep{cui2025entropy, hao2025rethinking, jin2025revisiting, yue2025does}. This exploration-exploitation imbalance \citep{chen2025exploration, huang2025beyond} has motivated various intervention strategies, including diversity-aware objectives \citep{yao2025diversity, he2025rewarding, walder2025pass, tang2025optimizing, yu2025dapo, gai2025differential, chen2025pass}, uncertainty-aware exploration \citep{xie2025unlocking}, and entropy control \citep{hao2025rethinking, park2025clip,cheng2025reasoning, agarwal2025unreasonable,li2025cure, wang2025beyond,zheng2025first,yang2025dcpo}. These techniques remain limited to local stochastic control in discrete space; whereas LaDi-RL instead performs exploration over entire reasoning trajectories in a continuous latent diffusion process, fundamentally altering the geometry of exploration. Due to the page limit, we discuss further related works in Appendix~\ref{app:additional_related_works}.
\vspace{-4mm}
\section{Conclusion}
\vspace{-3mm}
In this work, we propose LaDi-RL, a reinforcement learning method that reframes exploration in continuous latent space rather than relying on token-level sampling in discrete space. By modeling exploration as a multi-step latent diffusion process, LaDi-RL improves both pass@$1$ accuracy and pass@$k$ performance while mitigating diversity collapse. These results demonstrate that latent diffusion offers a principled and effective alternative to token-level RL for improving LLM reasoning.
\vspace{-3mm}


\bibliographystyle{plainnat}
\bibliography{citation}

@article{zhang2023planner,
  title={Planner: Generating diversified paragraph via latent language diffusion model},
  author={Zhang, Yizhe and Gu, Jiatao and Wu, Zhuofeng and Zhai, Shuangfei and Susskind, Joshua and Jaitly, Navdeep},
  journal={Advances in Neural Information Processing Systems},
  volume={36},
  pages={80178--80190},
  year={2023}
}

@article{meshchaninov2025compressed,
  title={Compressed and Smooth Latent Space for Text Diffusion Modeling},
  author={Meshchaninov, Viacheslav and Chimbulatov, Egor and Shabalin, Alexander and Abramov, Aleksandr and Vetrov, Dmitry},
  journal={arXiv preprint arXiv:2506.21170},
  year={2025}
}

@article{zeng2025treediff,
  title={TreeDiff: AST-Guided Code Generation with Diffusion LLMs},
  author={Zeng, Yiming and Cao, Jinghan and Li, Zexin and Chen, Yiming and Ren, Tao and Xiang, Dawei and Wu, Xidong and Gao, Shangqian and Yu, Tingting},
  journal={arXiv preprint arXiv:2508.01473},
  year={2025}
}

@inproceedings{singh2023codefusion,
  title={Codefusion: A pre-trained diffusion model for code generation},
  author={Singh, Mukul and Cambronero, Jos{\'e} and Gulwani, Sumit and Le, Vu and Negreanu, Carina and Verbruggen, Gust},
  booktitle={Proceedings of the 2023 Conference on Empirical Methods in Natural Language Processing},
  pages={11697--11708},
  year={2023}
}

@article{xiang2024diffusiondialog,
  title={Diffusiondialog: A diffusion model for diverse dialog generation with latent space},
  author={Xiang, Jianxiang and Liu, Zhenhua and Liu, Haodong and Bai, Yin and Cheng, Jia and Chen, Wenliang},
  journal={arXiv preprint arXiv:2404.06760},
  year={2024}
}

@article{gong2022diffuseq,
  title={Diffuseq: Sequence to sequence text generation with diffusion models},
  author={Gong, Shansan and Li, Mukai and Feng, Jiangtao and Wu, Zhiyong and Kong, LingPeng},
  journal={arXiv preprint arXiv:2210.08933},
  year={2022}
}

@article{li2022diffusion,
  title={Diffusion-lm improves controllable text generation},
  author={Li, Xiang and Thickstun, John and Gulrajani, Ishaan and Liang, Percy S and Hashimoto, Tatsunori B},
  journal={Advances in neural information processing systems},
  volume={35},
  pages={4328--4343},
  year={2022}
}

@inproceedings{lovelacestop,
  title={Stop-Think-AutoRegress: Language Modeling with Latent Diffusion Planning},
  author={Lovelace, Justin and Belardi, Christian K and Zalouk, Sofian and Polavaram, Adhitya and Kundurthy, Srivatsa R and Weinberger, Kilian Q},
  booktitle={Second Conference on Language Modeling}
}

@article{lovelace2023latent,
  title={Latent diffusion for language generation},
  author={Lovelace, Justin and Kishore, Varsha and Wan, Chao and Shekhtman, Eliot and Weinberger, Kilian Q},
  journal={Advances in Neural Information Processing Systems},
  volume={36},
  pages={56998--57025},
  year={2023}
}

@article{sahoo2024simple,
  title={Simple and effective masked diffusion language models},
  author={Sahoo, Subham and Arriola, Marianne and Schiff, Yair and Gokaslan, Aaron and Marroquin, Edgar and Chiu, Justin and Rush, Alexander and Kuleshov, Volodymyr},
  journal={Advances in Neural Information Processing Systems},
  volume={37},
  pages={130136--130184},
  year={2024}
}

@article{ho2020denoising,
  title={Denoising diffusion probabilistic models},
  author={Ho, Jonathan and Jain, Ajay and Abbeel, Pieter},
  journal={Advances in neural information processing systems},
  volume={33},
  pages={6840--6851},
  year={2020}
}

@article{ho2022classifier,
  title={Classifier-free diffusion guidance},
  author={Ho, Jonathan and Salimans, Tim},
  journal={arXiv preprint arXiv:2207.12598},
  year={2022}
}

@article{yue2025does,
  title={Does reinforcement learning really incentivize reasoning capacity in llms beyond the base model?},
  author={Yue, Yang and Chen, Zhiqi and Lu, Rui and Zhao, Andrew and Wang, Zhaokai and Song, Shiji and Huang, Gao},
  journal={arXiv preprint arXiv:2504.13837},
  year={2025}
}

@article{lipman2022flow,
  title={Flow matching for generative modeling},
  author={Lipman, Yaron and Chen, Ricky TQ and Ben-Hamu, Heli and Nickel, Maximilian and Le, Matt},
  journal={arXiv preprint arXiv:2210.02747},
  year={2022}
}

@article{chen2025pass,
  title={Pass@ k training for adaptively balancing exploration and exploitation of large reasoning models},
  author={Chen, Zhipeng and Qin, Xiaobo and Wu, Youbin and Ling, Yue and Ye, Qinghao and Zhao, Wayne Xin and Shi, Guang},
  journal={arXiv preprint arXiv:2508.10751},
  year={2025}
}

@article{song2020score,
  title={Score-based generative modeling through stochastic differential equations},
  author={Song, Yang and Sohl-Dickstein, Jascha and Kingma, Diederik P and Kumar, Abhishek and Ermon, Stefano and Poole, Ben},
  journal={arXiv preprint arXiv:2011.13456},
  year={2020}
}

@article{agarwal2025unreasonable,
  title={The unreasonable effectiveness of entropy minimization in llm reasoning},
  author={Agarwal, Shivam and Zhang, Zimin and Yuan, Lifan and Han, Jiawei and Peng, Hao},
  journal={arXiv preprint arXiv:2505.15134},
  year={2025}
}

@article{gai2025differential,
  title={Differential Smoothing Mitigates Sharpening and Improves LLM Reasoning},
  author={Gai, Jingchu and Zeng, Guanning and Zhang, Huaqing and Raghunathan, Aditi},
  journal={arXiv preprint arXiv:2511.19942},
  year={2025}
}

@article{tang2025optimizing,
  title={Optimizing language models for inference time objectives using reinforcement learning},
  author={Tang, Yunhao and Zheng, Kunhao and Synnaeve, Gabriel and Munos, R{\'e}mi},
  journal={arXiv preprint arXiv:2503.19595},
  year={2025}
}

@article{walder2025pass,
  title={Pass@ K Policy Optimization: Solving Harder Reinforcement Learning Problems},
  author={Walder, Christian and Karkhanis, Deep},
  journal={arXiv preprint arXiv:2505.15201},
  year={2025}
}

@article{ye2025dream,
  title={Dream 7B: Diffusion Large Language Models},
  author={Ye, Jiacheng and Xie, Zhihui and Zheng, Lin and Gao, Jiahui and Wu, Zirui and Jiang, Xin and Li, Zhenguo and Kong, Lingpeng},
  journal={arXiv preprint arXiv:2508.15487},
  year={2025}
}

@article{liu2025flow,
  title={Flow-grpo: Training flow matching models via online rl},
  author={Liu, Jie and Liu, Gongye and Liang, Jiajun and Li, Yangguang and Liu, Jiaheng and Wang, Xintao and Wan, Pengfei and Zhang, Di and Ouyang, Wanli},
  journal={arXiv preprint arXiv:2505.05470},
  year={2025}
}

@article{he2024olympiadbench,
  title={Olympiadbench: A challenging benchmark for promoting agi with olympiad-level bilingual multimodal scientific problems},
  author={He, Chaoqun and Luo, Renjie and Bai, Yuzhuo and Hu, Shengding and Thai, Zhen Leng and Shen, Junhao and Hu, Jinyi and Han, Xu and Huang, Yujie and Zhang, Yuxiang and others},
  journal={arXiv preprint arXiv:2402.14008},
  year={2024}
}

@article{hendrycks2021measuring,
  title={Measuring mathematical problem solving with the math dataset},
  author={Hendrycks, Dan and Burns, Collin and Kadavath, Saurav and Arora, Akul and Basart, Steven and Tang, Eric and Song, Dawn and Steinhardt, Jacob},
  journal={arXiv preprint arXiv:2103.03874},
  year={2021}
}

@article{chung2025revisiting,
  title={Revisiting Test-Time Scaling: A Survey and a Diversity-Aware Method for Efficient Reasoning},
  author={Chung, Ho-Lam and Hsiao, Teng-Yun and Huang, Hsiao-Ying and Cho, Chunerh and Lin, Jian-Ren and Ziwei, Zhang and Chen, Yun-Nung},
  journal={arXiv preprint arXiv:2506.04611},
  year={2025}
}

@article{chen2025rethinking,
  title={Rethinking fine-tuning when scaling test-time compute: Limiting confidence improves mathematical reasoning},
  author={Chen, Feng and Raventos, Allan and Cheng, Nan and Ganguli, Surya and Druckmann, Shaul},
  journal={arXiv preprint arXiv:2502.07154},
  year={2025}
}

@article{cobbe2021training,
  title={Training verifiers to solve math word problems},
  author={Cobbe, Karl and Kosaraju, Vineet and Bavarian, Mohammad and Chen, Mark and Jun, Heewoo and Kaiser, Lukasz and Plappert, Matthias and Tworek, Jerry and Hilton, Jacob and Nakano, Reiichiro and others},
  journal={arXiv preprint arXiv:2110.14168},
  year={2021}
}

@inproceedings{he2025rewarding,
  title={Rewarding the unlikely: Lifting grpo beyond distribution sharpening},
  author={He, Andre Wang and Fried, Daniel and Welleck, Sean},
  booktitle={Proceedings of the 2025 Conference on Empirical Methods in Natural Language Processing},
  pages={25559--25571},
  year={2025}
}

@article{zhao2025echo,
  title={Echo chamber: Rl post-training amplifies behaviors learned in pretraining},
  author={Zhao, Rosie and Meterez, Alexandru and Kakade, Sham and Pehlevan, Cengiz and Jelassi, Samy and Malach, Eran},
  journal={arXiv preprint arXiv:2504.07912},
  year={2025}
}

@article{dang2025weight,
  title={Weight ensembling improves reasoning in language models},
  author={Dang, Xingyu and Baek, Christina and Wen, Kaiyue and Kolter, Zico and Raghunathan, Aditi},
  journal={arXiv preprint arXiv:2504.10478},
  year={2025}
}

@article{song2024mind,
  title={Mind the gap: Examining the self-improvement capabilities of large language models},
  author={Song, Yuda and Zhang, Hanlin and Eisenach, Carson and Kakade, Sham and Foster, Dean and Ghai, Udaya},
  journal={arXiv preprint arXiv:2412.02674},
  year={2024}
}

@article{shao2024deepseekmath,
  title={Deepseekmath: Pushing the limits of mathematical reasoning in open language models},
  author={Shao, Zhihong and Wang, Peiyi and Zhu, Qihao and Xu, Runxin and Song, Junxiao and Bi, Xiao and Zhang, Haowei and Zhang, Mingchuan and Li, YK and Wu, Yang and others},
  journal={arXiv preprint arXiv:2402.03300},
  year={2024}
}

@misc{pfau2024letsthinkdotdot,
      title={Let's Think Dot by Dot: Hidden Computation in Transformer Language Models}, 
      author={Jacob Pfau and William Merrill and Samuel R. Bowman},
      year={2024},
      eprint={2404.15758},
      archivePrefix={arXiv},
      primaryClass={cs.CL},
      url={https://arxiv.org/abs/2404.15758}, 
}

@misc{herel2024thinkingtokenslanguagemodeling,
      title={Thinking Tokens for Language Modeling}, 
      author={David Herel and Tomas Mikolov},
      year={2024},
      eprint={2405.08644},
      archivePrefix={arXiv},
      primaryClass={cs.CL},
      url={https://arxiv.org/abs/2405.08644}, 
}

@inproceedings{
wang2024guiding,
title={Guiding Language Model Reasoning with Planning Tokens},
author={Xinyi Wang and Lucas Caccia and Oleksiy Ostapenko and Xingdi Yuan and William Yang Wang and Alessandro Sordoni},
booktitle={First Conference on Language Modeling},
year={2024},
url={https://openreview.net/forum?id=wi9IffRhVM}
}

@misc{jin2025disentanglingmemoryreasoningability,
      title={Disentangling Memory and Reasoning Ability in Large Language Models}, 
      author={Mingyu Jin and Weidi Luo and Sitao Cheng and Xinyi Wang and Wenyue Hua and Ruixiang Tang and William Yang Wang and Yongfeng Zhang},
      year={2025},
      eprint={2411.13504},
      archivePrefix={arXiv},
      primaryClass={cs.CL},
      url={https://arxiv.org/abs/2411.13504}, 
}

@inproceedings{
zelikman2024quietstar,
title={Quiet-{ST}aR: Language Models Can Teach Themselves to Think Before Speaking},
author={Eric Zelikman and Georges Raif Harik and Yijia Shao and Varuna Jayasiri and Nick Haber and Noah Goodman},
booktitle={First Conference on Language Modeling},
year={2024},
url={https://openreview.net/forum?id=oRXPiSOGH9}
}

@misc{hao2024traininglargelanguagemodels,
      title={Training Large Language Models to Reason in a Continuous Latent Space}, 
      author={Shibo Hao and Sainbayar Sukhbaatar and DiJia Su and Xian Li and Zhiting Hu and Jason Weston and Yuandong Tian},
      year={2024},
      eprint={2412.06769},
      archivePrefix={arXiv},
      primaryClass={cs.CL},
      url={https://arxiv.org/abs/2412.06769}, 
}

@misc{shen2025codicompressingchainofthoughtcontinuous,
      title={CODI: Compressing Chain-of-Thought into Continuous Space via Self-Distillation}, 
      author={Zhenyi Shen and Hanqi Yan and Linhai Zhang and Zhanghao Hu and Yali Du and Yulan He},
      year={2025},
      eprint={2502.21074},
      archivePrefix={arXiv},
      primaryClass={cs.CL},
      url={https://arxiv.org/abs/2502.21074}, 
}

@misc{liu2024expeditingelevatinglargelanguage,
      title={Expediting and Elevating Large Language Model Reasoning via Hidden Chain-of-Thought Decoding}, 
      author={Tianqiao Liu and Zui Chen and Zitao Liu and Mi Tian and Weiqi Luo},
      year={2024},
      eprint={2409.08561},
      archivePrefix={arXiv},
      primaryClass={cs.CL},
      url={https://arxiv.org/abs/2409.08561}, 
}

@misc{cheng2024compressedchainthoughtefficient,
      title={Compressed Chain of Thought: Efficient Reasoning Through Dense Representations}, 
      author={Jeffrey Cheng and Benjamin Van Durme},
      year={2024},
      eprint={2412.13171},
      archivePrefix={arXiv},
      primaryClass={cs.CL},
      url={https://arxiv.org/abs/2412.13171}, 
}

@misc{tack2025llmpretrainingcontinuousconcepts,
      title={LLM Pretraining with Continuous Concepts}, 
      author={Jihoon Tack and Jack Lanchantin and Jane Yu and Andrew Cohen and Ilia Kulikov and Janice Lan and Shibo Hao and Yuandong Tian and Jason Weston and Xian Li},
      year={2025},
      eprint={2502.08524},
      archivePrefix={arXiv},
      primaryClass={cs.LG},
      url={https://arxiv.org/abs/2502.08524}, 
}

@inproceedings{
mohtashami2025cotformer,
title={Co{TF}ormer: A Chain of Thought Driven Architecture with Budget-Adaptive Computation Cost at Inference},
author={Amirkeivan Mohtashami and Matteo Pagliardini and Martin Jaggi},
booktitle={The Thirteenth International Conference on Learning Representations},
year={2025},
url={https://openreview.net/forum?id=7igPXQFupX}
}

@misc{geiping2025scalingtesttimecomputelatent,
      title={Scaling up Test-Time Compute with Latent Reasoning: A Recurrent Depth Approach}, 
      author={Jonas Geiping and Sean McLeish and Neel Jain and John Kirchenbauer and Siddharth Singh and Brian R. Bartoldson and Bhavya Kailkhura and Abhinav Bhatele and Tom Goldstein},
      year={2025},
      eprint={2502.05171},
      archivePrefix={arXiv},
      primaryClass={cs.LG},
      url={https://arxiv.org/abs/2502.05171}, 
}

@misc{yu2025enhancingautoregressivechainofthoughtloopaligned,
      title={Enhancing Auto-regressive Chain-of-Thought through Loop-Aligned Reasoning}, 
      author={Qifan Yu and Zhenyu He and Sijie Li and Xun Zhou and Jun Zhang and Jingjing Xu and Di He},
      year={2025},
      eprint={2502.08482},
      archivePrefix={arXiv},
      primaryClass={cs.CL},
      url={https://arxiv.org/abs/2502.08482}, 
}

@misc{chen2025innerthinkingtransformerleveraging,
      title={Inner Thinking Transformer: Leveraging Dynamic Depth Scaling to Foster Adaptive Internal Thinking}, 
      author={Yilong Chen and Junyuan Shang and Zhenyu Zhang and Yanxi Xie and Jiawei Sheng and Tingwen Liu and Shuohuan Wang and Yu Sun and Hua Wu and Haifeng Wang},
      year={2025},
      eprint={2502.13842},
      archivePrefix={arXiv},
      primaryClass={cs.CL},
      url={https://arxiv.org/abs/2502.13842}, 
}

@misc{saunshi2025reasoninglatentthoughtspower,
      title={Reasoning with Latent Thoughts: On the Power of Looped Transformers}, 
      author={Nikunj Saunshi and Nishanth Dikkala and Zhiyuan Li and Sanjiv Kumar and Sashank J. Reddi},
      year={2025},
      eprint={2502.17416},
      archivePrefix={arXiv},
      primaryClass={cs.CL},
      url={https://arxiv.org/abs/2502.17416}, 
}

@misc{zhou2024transfusionpredicttokendiffuse,
      title={Transfusion: Predict the Next Token and Diffuse Images with One Multi-Modal Model}, 
      author={Chunting Zhou and Lili Yu and Arun Babu and Kushal Tirumala and Michihiro Yasunaga and Leonid Shamis and Jacob Kahn and Xuezhe Ma and Luke Zettlemoyer and Omer Levy},
      year={2024},
      eprint={2408.11039},
      archivePrefix={arXiv},
      primaryClass={cs.AI},
      url={https://arxiv.org/abs/2408.11039}, 
}

@misc{tong2024metamorphmultimodalunderstandinggeneration,
      title={MetaMorph: Multimodal Understanding and Generation via Instruction Tuning}, 
      author={Shengbang Tong and David Fan and Jiachen Zhu and Yunyang Xiong and Xinlei Chen and Koustuv Sinha and Michael Rabbat and Yann LeCun and Saining Xie and Zhuang Liu},
      year={2024},
      eprint={2412.14164},
      archivePrefix={arXiv},
      primaryClass={cs.CV},
      url={https://arxiv.org/abs/2412.14164}, 
}

@misc{fan2024fluidscalingautoregressivetexttoimage,
      title={Fluid: Scaling Autoregressive Text-to-image Generative Models with Continuous Tokens}, 
      author={Lijie Fan and Tianhong Li and Siyang Qin and Yuanzhen Li and Chen Sun and Michael Rubinstein and Deqing Sun and Kaiming He and Yonglong Tian},
      year={2024},
      eprint={2410.13863},
      archivePrefix={arXiv},
      primaryClass={cs.CV},
      url={https://arxiv.org/abs/2410.13863}, 
}

@misc{xiao2024omnigenunifiedimagegeneration,
      title={OmniGen: Unified Image Generation}, 
      author={Shitao Xiao and Yueze Wang and Junjie Zhou and Huaying Yuan and Xingrun Xing and Ruiran Yan and Chaofan Li and Shuting Wang and Tiejun Huang and Zheng Liu},
      year={2024},
      eprint={2409.11340},
      archivePrefix={arXiv},
      primaryClass={cs.CV},
      url={https://arxiv.org/abs/2409.11340}, 
}

@misc{black2024pi0visionlanguageactionflowmodel,
      title={$\pi_0$: A Vision-Language-Action Flow Model for General Robot Control}, 
      author={Kevin Black and Noah Brown and Danny Driess and Adnan Esmail and Michael Equi and Chelsea Finn and Niccolo Fusai and Lachy Groom and Karol Hausman and Brian Ichter and Szymon Jakubczak and Tim Jones and Liyiming Ke and Sergey Levine and Adrian Li-Bell and Mohith Mothukuri and Suraj Nair and Karl Pertsch and Lucy Xiaoyang Shi and James Tanner and Quan Vuong and Anna Walling and Haohuan Wang and Ury Zhilinsky},
      year={2024},
      eprint={2410.24164},
      archivePrefix={arXiv},
      primaryClass={cs.LG},
      url={https://arxiv.org/abs/2410.24164}, 
}

@misc{chen2025janusprounifiedmultimodalunderstanding,
      title={Janus-Pro: Unified Multimodal Understanding and Generation with Data and Model Scaling}, 
      author={Xiaokang Chen and Zhiyu Wu and Xingchao Liu and Zizheng Pan and Wen Liu and Zhenda Xie and Xingkai Yu and Chong Ruan},
      year={2025},
      eprint={2501.17811},
      archivePrefix={arXiv},
      primaryClass={cs.AI},
      url={https://arxiv.org/abs/2501.17811}, 
}

@misc{chen2025gokuflowbasedvideo,
      title={Goku: Flow Based Video Generative Foundation Models}, 
      author={Shoufa Chen and Chongjian Ge and Yuqi Zhang and Yida Zhang and Fengda Zhu and Hao Yang and Hongxiang Hao and Hui Wu and Zhichao Lai and Yifei Hu and Ting-Che Lin and Shilong Zhang and Fu Li and Chuan Li and Xing Wang and Yanghua Peng and Peize Sun and Ping Luo and Yi Jiang and Zehuan Yuan and Bingyue Peng and Xiaobing Liu},
      year={2025},
      eprint={2502.04896},
      archivePrefix={arXiv},
      primaryClass={cs.CV},
      url={https://arxiv.org/abs/2502.04896}, 
}

@misc{tang2024hartefficientvisualgeneration,
      title={HART: Efficient Visual Generation with Hybrid Autoregressive Transformer}, 
      author={Haotian Tang and Yecheng Wu and Shang Yang and Enze Xie and Junsong Chen and Junyu Chen and Zhuoyang Zhang and Han Cai and Yao Lu and Song Han},
      year={2024},
      eprint={2410.10812},
      archivePrefix={arXiv},
      primaryClass={cs.CV},
      url={https://arxiv.org/abs/2410.10812}, 
}

@misc{shi2025lmfusionadaptingpretrainedlanguage,
      title={LMFusion: Adapting Pretrained Language Models for Multimodal Generation}, 
      author={Weijia Shi and Xiaochuang Han and Chunting Zhou and Weixin Liang and Xi Victoria Lin and Luke Zettlemoyer and Lili Yu},
      year={2025},
      eprint={2412.15188},
      archivePrefix={arXiv},
      primaryClass={cs.CL},
      url={https://arxiv.org/abs/2412.15188}, 
}

@misc{pan2025transfermodalitiesmetaqueries,
      title={Transfer between Modalities with MetaQueries}, 
      author={Xichen Pan and Satya Narayan Shukla and Aashu Singh and Zhuokai Zhao and Shlok Kumar Mishra and Jialiang Wang and Zhiyang Xu and Jiuhai Chen and Kunpeng Li and Felix Juefei-Xu and Ji Hou and Saining Xie},
      year={2025},
      eprint={2504.06256},
      archivePrefix={arXiv},
      primaryClass={cs.CV},
      url={https://arxiv.org/abs/2504.06256}, 
}

@article{wu2025llms,
  title={LLMs are Single-threaded Reasoners: Demystifying the Working Mechanism of Soft Thinking},
  author={Wu, Ch{\"u}nhung and Lu, Jinliang and Ren, Zixuan and Hu, Gangqiang and Wu, Zhi and Dai, Dai and Wu, Hua},
  journal={arXiv preprint arXiv:2508.03440},
  year={2025}
}

@article{zhu2025reasoning,
  title={Reasoning by Superposition: A Theoretical Perspective on Chain of Continuous Thought},
  author={Zhu, Hanlin and Hao, Shibo and Hu, Zhiting and Jiao, Jiantao and Russell, Stuart and Tian, Yuandong},
  journal={arXiv preprint arXiv:2505.12514},
  year={2025}
}

@article{zhu2025scaling,
  title={Scaling latent reasoning via looped language models},
  author={Zhu, Rui-Jie and Wang, Zixuan and Hua, Kai and Zhang, Tianyu and Li, Ziniu and Que, Haoran and Wei, Boyi and Wen, Zixin and Yin, Fan and Xing, He and others},
  journal={arXiv preprint arXiv:2510.25741},
  year={2025}
}

@article{fu2025think,
  title={Think-at-Hard: Selective Latent Iterations to Improve Reasoning Language Models},
  author={Fu, Tianyu and You, Yichen and Chen, Zekai and Dai, Guohao and Yang, Huazhong and Wang, Yu},
  journal={arXiv preprint arXiv:2511.08577},
  year={2025}
}

@article{yang2025qwen3,
  title={Qwen3 technical report},
  author={Yang, An and Li, Anfeng and Yang, Baosong and Zhang, Beichen and Hui, Binyuan and Zheng, Bo and Yu, Bowen and Gao, Chang and Huang, Chengen and Lv, Chenxu and others},
  journal={arXiv preprint arXiv:2505.09388},
  year={2025}
}

@inproceedings{huang2025opencoder,
  title={Opencoder: The open cookbook for top-tier code large language models},
  author={Huang, Siming and Cheng, Tianhao and Liu, Jason Klein and Xu, Weidi and Hao, Jiaran and Song, Liuyihan and Xu, Yang and Yang, Jian and Liu, Jiaheng and Zhang, Chenchen and others},
  booktitle={Proceedings of the 63rd Annual Meeting of the Association for Computational Linguistics (Volume 1: Long Papers)},
  pages={33167--33193},
  year={2025}
}

@article{hui2024qwen2,
  title={Qwen2. 5-coder technical report},
  author={Hui, Binyuan and Yang, Jian and Cui, Zeyu and Yang, Jiaxi and Liu, Dayiheng and Zhang, Lei and Liu, Tianyu and Zhang, Jiajun and Yu, Bowen and Lu, Keming and others},
  journal={arXiv preprint arXiv:2409.12186},
  year={2024}
}

@misc{codefuse2025samplemattersleveragingmixtureofexperts,
      title={Every Sample Matters: Leveraging Mixture-of-Experts and High-Quality Data for Efficient and Accurate Code LLM}, 
      author={Codefuse and Ling Team},
      year={2025},
      eprint={2503.17793},
      archivePrefix={arXiv},
      primaryClass={cs.LG},
      url={https://arxiv.org/abs/2503.17793}, 
}

@article{butt2025soft,
  title={Soft Tokens, Hard Truths},
  author={Butt, Natasha and Kwiatkowski, Ariel and Labiad, Ismail and Kempe, Julia and Ollivier, Yann},
  journal={arXiv preprint arXiv:2509.19170},
  year={2025}
}

@article{naik2023diversity,
  title={Diversity of thought improves reasoning abilities of llms},
  author={Naik, Ranjita and Chandrasekaran, Varun and Yuksekgonul, Mert and Palangi, Hamid and Nushi, Besmira},
  journal={arXiv preprint arXiv:2310.07088},
  year={2023}
}

@article{zhang2025soft,
  title={Soft thinking: Unlocking the reasoning potential of llms in continuous concept space},
  author={Zhang, Zhen and He, Xuehai and Yan, Weixiang and Shen, Ao and Zhao, Chenyang and Wang, Shuohang and Shen, Yelong and Wang, Xin Eric},
  journal={arXiv preprint arXiv:2505.15778},
  year={2025}
}

@article{guo2025deepseek,
  title={DeepSeek-R1: Incentivizing Reasoning Capability in LLMs via Reinforcement Learning},
  author={Guo, Daya and Yang, Dejian and Zhang, Haowei and Song, Junxiao and Wang, Peiyi and Zhu, Qihao and Xu, Runxin and Zhang, Ruoyu and Ma, Shirong and Bi, Xiao and others},
  journal={arXiv preprint arXiv:2501.12948},
  year={2025}
}

@article{deepseek2024v3,
  title={DeepSeek-V3 Technical Report},
  author={DeepSeek-AI},
  journal={arXiv preprint arXiv:2412.19437},
  year={2024}
}

@article{openai2024o1,
  title={Learning to Reason with LLMs},
  author={OpenAI},
  journal={OpenAI Blog},
  year={2024}
}

@article{schulman2017ppo,
  title={Proximal Policy Optimization Algorithms},
  author={Schulman, John and Wolski, Filip and Dhariwal, Prafulla and Radford, Alec and Klimov, Oleg},
  journal={arXiv preprint arXiv:1707.06347},
  year={2017}
}

@article{rafailov2023dpo,
  title={Direct Preference Optimization: Your Language Model is Secretly a Reward Model},
  author={Rafailov, Rafael and Sharma, Archit and Mitchell, Eric and Ermon, Stefano and Manning, Christopher D and Finn, Chelsea},
  journal={arXiv preprint arXiv:2305.18290},
  year={2023}
}

@article{yu2025dapo,
  title={DAPO: An Open-Source LLM Reinforcement Learning System at Scale},
  author={Yu, Qiying and Zhang, Zheng and Zhu, Ruofei and Yuan, Yufeng and Zuo, Xiaochen and others},
  journal={arXiv preprint arXiv:2503.14476},
  year={2025}
}

@article{lin2025cppo,
  title={CPPO: Accelerating the Training of Group Relative Policy Optimization-based Reasoning Models},
  author={Lin, Z and Lin, M and Xie, Y and Ji, R},
  journal={arXiv preprint arXiv:2503.22342},
  year={2025}
}

@article{zhang2025stepwise,
  title={R1-VL: Learning to Reason with Multimodal Large Language Models via Step-wise Group Relative Policy Optimization},
  author={Zhang, J and Huang, J and Yao, H and Liu, S and Zhang, X and Lu, S and others},
  journal={arXiv preprint arXiv:2503.12937},
  year={2025}
}

@article{li2025adaptive,
  title={Adaptive Group Policy Optimization: Towards Stable Training and Token-Efficient Reasoning},
  author={Li, C and Liu, N and Yang, K},
  journal={arXiv preprint arXiv:2503.15952},
  year={2025}
}

@article{cheng2025reasoning,
  title={Reasoning with exploration: An entropy perspective},
  author={Cheng, Daixuan and Huang, Shaohan and Zhu, Xuekai and Dai, Bo and Zhao, Wayne Xin and Zhang, Zhenliang and Wei, Furu},
  journal={arXiv preprint arXiv:2506.14758},
  year={2025}
}

@article{hao2025rethinking,
  title={Rethinking Entropy Interventions in RLVR: An Entropy Change Perspective},
  author={Hao, Zhezheng and Wang, Hong and Liu, Haoyang and Luo, Jian and Yu, Jiarui and others},
  journal={arXiv preprint arXiv:2510.10150},
  year={2025}
}

@article{jin2025revisiting,
  title={Revisiting Entropy in Reinforcement Learning for Large Reasoning Models},
  author={Jin, R and Gao, Y and others},
  journal={arXiv preprint arXiv:2511.05993},
  year={2025}
}

@article{park2025clip,
  title={Clip-Low Increases Entropy and Clip-High Decreases Entropy in Reinforcement Learning of Large Language Models},
  author={Park, JR and others},
  journal={arXiv preprint arXiv:2509.26114},
  year={2025}
}

@article{yao2025diversity,
  title={Diversity-Aware Policy Optimization for Large Language Model Reasoning},
  author={Yao, Jian and Cheng, Ran and Wu, Xingyu and Wu, Jibin and Tan, Kay Chen},
  journal={arXiv preprint arXiv:2505.23433},
  year={2025}
}

@article{zheng2025first,
  title={First return, entropy-eliciting explore},
  author={Zheng, Tianyu and Xing, Tianshun and Gu, Qingshui and Liang, Taoran and Qu, Xingwei and Zhou, Xin and Li, Yizhi and Wen, Zhoufutu and Lin, Chenghua and Huang, Wenhao and others},
  journal={arXiv preprint arXiv:2507.07017},
  year={2025}
}

@article{yang2025dcpo,
  title={Dcpo: Dynamic clipping policy optimization},
  author={Yang, Shihui and Dou, Chengfeng and Guo, Peidong and Lu, Kai and Ju, Qiang and Deng, Fei and Xin, Rihui},
  journal={arXiv preprint arXiv:2509.02333},
  year={2025}
}

@article{chen2025exploration,
  title={Exploration vs Exploitation: Rethinking RLVR through Clipping, Entropy, and Spurious Reward},
  author={Chen, P and others},
  journal={arXiv preprint arXiv:2512.16912},
  year={2025}
}

@article{wang2025beyond,
  title={Beyond the 80/20 rule: High-entropy minority tokens drive effective reinforcement learning for llm reasoning},
  author={Wang, Shenzhi and Yu, Le and Gao, Chang and Zheng, Chujie and Liu, Shixuan and Lu, Rui and Dang, Kai and Chen, Xionghui and Yang, Jianxin and Zhang, Zhenru and others},
  journal={arXiv preprint arXiv:2506.01939},
  year={2025}
}

@article{huang2025beyond,
  title={Beyond the Exploration-Exploitation Trade-off: A Hidden State Approach for LLM Reasoning in RLVR},
  author={Huang, F and others},
  journal={arXiv preprint arXiv:2509.23808},
  year={2025}
}

@article{xie2025unlocking,
  title={Unlocking Exploration in RLVR: Uncertainty-aware Exploration Strategy},
  author={Xie, C and others},
  journal={arXiv preprint arXiv:2509.06941},
  year={2025}
}

@article{hu2025diversity,
  title={Diversity-incentivized exploration for versatile reasoning},
  author={Hu, Zican and Zhang, Shilin and Li, Yafu and Yan, Jianhao and Hu, Xuyang and Cui, Leyang and Qu, Xiaoye and Chen, Chunlin and Cheng, Yu and Wang, Zhi},
  journal={arXiv preprint arXiv:2509.26209},
  year={2025}
}

@article{zheng2025soft,
  title={SofT-GRPO: Surpassing Discrete-Token LLM Reinforcement Learning via Gumbel-Reparameterized Soft-Thinking Policy Optimization},
  author={Zheng, Zhi and Lee, Wee Sun},
  journal={arXiv preprint arXiv:2511.06411},
  year={2025}
}

@article{tang2026multiplex,
  title={Multiplex Thinking: Reasoning via Token-wise Branch-and-Merge},
  author={Tang, Yao and Dong, Li and Hao, Yaru and Dong, Qingxiu and Wei, Furu and Gu, Jiatao},
  journal={arXiv preprint arXiv:2601.08808},
  year={2026}
}

@article{xu2025kodcode,
  title={Kodcode: A diverse, challenging, and verifiable synthetic dataset for coding},
  author={Xu, Zhangchen and Liu, Yang and Yin, Yueqin and Zhou, Mingyuan and Poovendran, Radha},
  journal={arXiv preprint arXiv:2503.02951},
  year={2025}
}

@article{AceCoder,
    title={AceCoder: Acing Coder RL via Automated Test-Case Synthesis},
    author={Zeng, Huaye and Jiang, Dongfu and Wang, Haozhe and Nie, Ping and Chen, Xiaotong and Chen, Wenhu},
    journal={ArXiv},
    year={2025},
    volume={abs/2207.01780}
}

@article{lewkowycz2022solving,
  title={Solving quantitative reasoning problems with language models},
  author={Lewkowycz, Aitor and Andreassen, Anders and Dohan, David and Dyer, Ethan and Michalewski, Henryk and Ramasesh, Vinay and Slone, Ambrose and Anil, Cem and Schlag, Imanol and Gutman-Solo, Theo and others},
  journal={Advances in neural information processing systems},
  volume={35},
  pages={3843--3857},
  year={2022}
}

@misc{aime25,
      title={American Invitational Mathematics Examination (AIME) 2025}, 
      author={Zhang, Yifan and Math-AI, Team},
      year={2025},
}

@dataset{aime_1983_2024,
  author = {Hemish Veeraboina},
  title = {AIME Problem Set 1983-2024},
  year = {2023},
  publisher = {Kaggle},
  url = {https://www.kaggle.com/datasets/hemishveeraboina/aime-problem-set-1983-2024}
}

@misc{slam-distillation-from-r1, 
    author = {Sathwik Tejaswi Madhusudhan and Shruthan Radhakrishna and Jash Mehta and Toby Liang},  
    title = {Millions scale dataset distilled from R1-32b},  
    howpublished = {https://huggingface.co/datasets/ServiceNow-AI/R1-Distill-SFT},
    publisher = {SLAM - ServiceNow Language Models Lab}  
}

@article{wang2025d2,
  title={d2: Improved techniques for training reasoning diffusion language models},
  author={Wang, Guanghan and Schiff, Yair and Turok, Gilad and Kuleshov, Volodymyr},
  journal={arXiv preprint arXiv:2509.21474},
  year={2025}
}

@inproceedings{lovelace2024diffusion,
  title={Diffusion guided language modeling},
  author={Lovelace, Justin and Kishore, Varsha and Chen, Yiwei and Weinberger, Kilian},
  booktitle={Findings of the Association for Computational Linguistics: ACL 2024},
  pages={14936--14952},
  year={2024}
}

@article{d2021repulsive,
  title={Repulsive deep ensembles are bayesian},
  author={D'Angelo, Francesco and Fortuin, Vincent},
  journal={Advances in Neural Information Processing Systems},
  volume={34},
  pages={3451--3465},
  year={2021}
}

@article{zilberstein2024repulsive,
  title={Repulsive latent score distillation for solving inverse problems},
  author={Zilberstein, Nicolas and Mardani, Morteza and Segarra, Santiago},
  journal={arXiv preprint arXiv:2406.16683},
  year={2024}
}

@article{wang2025revolutionizing,
  title={Revolutionizing reinforcement learning framework for diffusion large language models},
  author={Wang, Yinjie and Yang, Ling and Li, Bowen and Tian, Ye and Shen, Ke and Wang, Mengdi},
  journal={arXiv preprint arXiv:2509.06949},
  year={2025}
}

@article{huang2025reinforcing,
  title={Reinforcing the diffusion chain of lateral thought with diffusion language models},
  author={Huang, Zemin and Chen, Zhiyang and Wang, Zijun and Li, Tiancheng and Qi, Guo-Jun},
  journal={arXiv preprint arXiv:2505.10446},
  year={2025}
}

@article{tang2025wd1,
  title={wd1: Weighted policy optimization for reasoning in diffusion language models},
  author={Tang, Xiaohang and Dolga, Rares and Yoon, Sangwoong and Bogunovic, Ilija},
  journal={arXiv preprint arXiv:2507.08838},
  year={2025}
}

@inproceedings{schulman2015trust,
  title={Trust region policy optimization},
  author={Schulman, John and Levine, Sergey and Abbeel, Pieter and Jordan, Michael and Moritz, Philipp},
  booktitle={International conference on machine learning},
  pages={1889--1897},
  year={2015},
  organization={PMLR}
}

@article{rojas2025improving,
  title={Improving reasoning for diffusion language models via group diffusion policy optimization},
  author={Rojas, Kevin and Lin, Jiahe and Rasul, Kashif and Schneider, Anderson and Nevmyvaka, Yuriy and Tao, Molei and Deng, Wei},
  journal={arXiv preprint arXiv:2510.08554},
  year={2025}
}

@article{ou2025principled,
  title={Principled rl for diffusion llms emerges from a sequence-level perspective},
  author={Ou, Jingyang and Han, Jiaqi and Xu, Minkai and Xu, Shaoxuan and Xie, Jianwen and Ermon, Stefano and Wu, Yi and Li, Chongxuan},
  journal={arXiv preprint arXiv:2512.03759},
  year={2025}
}

@article{wang2025spg,
  title={Spg: Sandwiched policy gradient for masked diffusion language models},
  author={Wang, Chenyu and Rashidinejad, Paria and Su, DiJia and Jiang, Song and Wang, Sid and Zhao, Siyan and Zhou, Cai and Shen, Shannon Zejiang and Chen, Feiyu and Jaakkola, Tommi and others},
  journal={arXiv preprint arXiv:2510.09541},
  year={2025}
}

@article{zhu2025llada,
  title={LLaDA 1.5: Variance-Reduced Preference Optimization for Large Language Diffusion Models},
  author={Zhu, Fengqi and Wang, Rongzhen and Nie, Shen and Zhang, Xiaolu and Wu, Chunwei and Hu, Jun and Zhou, Jun and Chen, Jianfei and Lin, Yankai and Wen, Ji-Rong and others},
  journal={arXiv preprint arXiv:2505.19223},
  year={2025}
}

@article{yang2025mmada,
  title={Mmada: Multimodal large diffusion language models},
  author={Yang, Ling and Tian, Ye and Li, Bowen and Zhang, Xinchen and Shen, Ke and Tong, Yunhai and Wang, Mengdi},
  journal={arXiv preprint arXiv:2505.15809},
  year={2025}
}

@article{sahoo2025diffusion,
  title={The diffusion duality},
  author={Sahoo, Subham Sekhar and Deschenaux, Justin and Gokaslan, Aaron and Wang, Guanghan and Chiu, Justin and Kuleshov, Volodymyr},
  journal={arXiv preprint arXiv:2506.10892},
  year={2025}
}

@article{zekri2025fine,
  title={Fine-tuning discrete diffusion models with policy gradient methods},
  author={Zekri, Oussama and Boull{\'e}, Nicolas},
  journal={arXiv preprint arXiv:2502.01384},
  year={2025}
}

@article{li2025mixgrpo,
  title={Mixgrpo: Unlocking flow-based grpo efficiency with mixed ode-sde},
  author={Li, Junzhe and Cui, Yutao and Huang, Tao and Ma, Yinping and Fan, Chun and Yang, Miles and Zhong, Zhao},
  journal={arXiv preprint arXiv:2507.21802},
  year={2025}
}

@article{xue2025dancegrpo,
  title={DanceGRPO: Unleashing GRPO on Visual Generation},
  author={Xue, Zeyue and Wu, Jie and Gao, Yu and Kong, Fangyuan and Zhu, Lingting and Chen, Mengzhao and Liu, Zhiheng and Liu, Wei and Guo, Qiushan and Huang, Weilin and others},
  journal={arXiv preprint arXiv:2505.07818},
  year={2025}
}

@article{borso2025d3po,
  title={D3PO: Preference-Based Alignment of Discrete Diffusion Models},
  author={Borso, Umberto and Paglieri, Davide and Wells, Jude and Rockt{\"a}schel, Tim},
  journal={arXiv e-prints},
  pages={arXiv--2503},
  year={2025}
}

@article{zhao2025diffpo,
  title={Diffpo: Training diffusion llms to reason fast and furious via reinforcement learning},
  author={Zhao, Hanyang and Liang, Dawen and Tang, Wenpin and Yao, David and Kallus, Nathan},
  journal={arXiv preprint arXiv:2510.02212},
  year={2025}
}

@article{song2025seed,
  title={Seed diffusion: A large-scale diffusion language model with high-speed inference},
  author={Song, Yuxuan and Zhang, Zheng and Luo, Cheng and Gao, Pengyang and Xia, Fan and Luo, Hao and Li, Zheng and Yang, Yuehang and Yu, Hongli and Qu, Xingwei and others},
  journal={arXiv preprint arXiv:2508.02193},
  year={2025}
}

@article{song2019generative,
  title={Generative modeling by estimating gradients of the data distribution},
  author={Song, Yang and Ermon, Stefano},
  journal={Advances in neural information processing systems},
  volume={32},
  year={2019}
}

@article{qiu2025noisygrpo,
  title={NoisyGRPO: Incentivizing Multimodal CoT Reasoning via Noise Injection and Bayesian Estimation},
  author={Qiu, Longtian and Ning, Shan and Sun, Jiaxuan and He, Xuming},
  journal={arXiv preprint arXiv:2510.21122},
  year={2025}
}

@misc{deepscaler2025,
  title={DeepScaleR: Surpassing O1-Preview with a 1.5B Model by Scaling RL},
  author={Michael Luo and Sijun Tan and Justin Wong and Xiaoxiang Shi and William Y. Tang and Manan Roongta and Colin Cai and Jeffrey Luo and Li Erran Li and Raluca Ada Popa and Ion Stoica},
  howpublished={\url{https://pretty-radio-b75.notion.site/DeepScaleR-Surpassing-O1-Preview-with-a-1-5B-Model-by-Scaling-RL-19681902c1468005bed8ca303013a4e2}},
  note={Notion Blog},
  year={2025}
}

@article{ozeren2025reinforcement,
  title={Reinforcement Learning for Latent-Space Thinking in LLMs},
  author={{\"O}zeren, Enes and A{\ss}enmacher, Matthias},
  journal={arXiv preprint arXiv:2512.11816},
  year={2025}
}

@article{li2025cure,
  title={Cure: Critical-token-guided re-concatenation for entropy-collapse prevention},
  author={Li, Qingbin and Xue, Rongkun and Wang, Jie and Zhou, Ming and Li, Zhi and Ji, Xiaofeng and Wang, Yongqi and Liu, Miao and Yang, Zheming and Qiu, Minghui and others},
  journal={arXiv preprint arXiv:2508.11016},
  year={2025}
}

@article{black2023training,
  title={Training diffusion models with reinforcement learning},
  author={Black, Kevin and Janner, Michael and Du, Yilun and Kostrikov, Ilya and Levine, Sergey},
  journal={arXiv preprint arXiv:2305.13301},
  year={2023}
}

@article{anon2024diffusion,
  title={Diffusion of Thought: Chain-of-Thought Reasoning in Diffusion Language Models},
  author={Anonymous},
  journal={NeurIPS},
  year={2024}
}

@article{shao2025diffuse,
  title={Diffuse Thinking: Exploring Diffusion Language Models as Efficient Thought Proposers for Reasoning},
  author={Shao, C and others},
  journal={arXiv preprint arXiv:2510.27469},
  year={2025}
}

@article{venkatraman2024latent,
  title={Reasoning with Latent Diffusion in Offline Reinforcement Learning},
  author={Venkatraman, S and others},
  journal={ICLR},
  year={2024}
}

@article{zhang2025survey,
  title={A Survey of Reinforcement Learning for Large Reasoning Models},
  author={Zhang, K and others},
  journal={arXiv preprint arXiv:2509.08827},
  year={2025}
}

@article{albergo2023stochastic,
  title={Stochastic interpolants: A unifying framework for flows and diffusions},
  author={Albergo, Michael S and Boffi, Nicholas M and Vanden-Eijnden, Eric},
  journal={arXiv preprint arXiv:2303.08797},
  year={2023}
}

@article{liu2022flow,
  title={Flow straight and fast: Learning to generate and transfer data with rectified flow},
  author={Liu, Xingchao and Gong, Chengyue and Liu, Qiang},
  journal={arXiv preprint arXiv:2209.03003},
  year={2022}
}

@article{zhang2025relax,
  title={ReLaX: Reasoning with Latent Exploration for Large Reasoning Models},
  author={Zhang, Shimin and Chen, Xianwei and Shen, Yufan and Ye, Ziyuan and Wu, Jibin},
  journal={arXiv preprint arXiv:2512.07558},
  year={2025}
}

@article{sane2025hybrid,
  title={Hybrid group relative policy optimization: A multi-sample approach to enhancing policy optimization},
  author={Sane, Soham},
  journal={arXiv preprint arXiv:2502.01652},
  year={2025}
}

@article{zhao2025d1,
  title={d1: Scaling reasoning in diffusion large language models via reinforcement learning},
  author={Zhao, Siyan and Gupta, Devaansh and Zheng, Qinqing and Grover, Aditya},
  journal={arXiv preprint arXiv:2504.12216},
  year={2025}
}

@article{nie2025large,
  title={Large language diffusion models},
  author={Nie, Shen and Zhu, Fengqi and You, Zebin and Zhang, Xiaolu and Ou, Jingyang and Hu, Jun and Zhou, Jun and Lin, Yankai and Wen, Ji-Rong and Li, Chongxuan},
  journal={arXiv preprint arXiv:2502.09992},
  year={2025}
}

@article{xie2025dream,
  title={Dream-coder 7b: An open diffusion language model for code},
  author={Xie, Zhihui and Ye, Jiacheng and Zheng, Lin and Gao, Jiahui and Dong, Jingwei and Wu, Zirui and Zhao, Xueliang and Gong, Shansan and Jiang, Xin and Li, Zhenguo and others},
  journal={arXiv preprint arXiv:2509.01142},
  year={2025}
}

@article{wang2025coefficients,
  title={Coefficients-Preserving Sampling for Reinforcement Learning with Flow Matching},
  author={Wang, Feng and Yu, Zihao},
  journal={arXiv preprint arXiv:2509.05952},
  year={2025}
}

@article{cui2025entropy,
  title={The entropy mechanism of reinforcement learning for reasoning language models},
  author={Cui, Ganqu and Zhang, Yuchen and Chen, Jiacheng and Yuan, Lifan and Wang, Zhi and Zuo, Yuxin and Li, Haozhan and Fan, Yuchen and Chen, Huayu and Chen, Weize and others},
  journal={arXiv preprint arXiv:2505.22617},
  year={2025}
}

@article{liu2023your,
  title={Is your code generated by chatgpt really correct? rigorous evaluation of large language models for code generation},
  author={Liu, Jiawei and Xia, Chunqiu Steven and Wang, Yuyao and Zhang, Lingming},
  journal={Advances in Neural Information Processing Systems},
  volume={36},
  pages={21558--21572},
  year={2023}
}

@article{jain2024livecodebench,
  title={Livecodebench: Holistic and contamination free evaluation of large language models for code},
  author={Jain, Naman and Han, King and Gu, Alex and Li, Wen-Ding and Yan, Fanjia and Zhang, Tianjun and Wang, Sida and Solar-Lezama, Armando and Sen, Koushik and Stoica, Ion},
  journal={arXiv preprint arXiv:2403.07974},
  year={2024}
}

@article{austin2021program,
  title={Program synthesis with large language models},
  author={Austin, Jacob and Odena, Augustus and Nye, Maxwell and Bosma, Maarten and Michalewski, Henryk and Dohan, David and Jiang, Ellen and Cai, Carrie and Terry, Michael and Le, Quoc and others},
  journal={arXiv preprint arXiv:2108.07732},
  year={2021}
}

@article{chen2021evaluating,
  title={Evaluating large language models trained on code},
  author={Chen, Mark},
  journal={arXiv preprint arXiv:2107.03374},
  year={2021}
}

@article{gozeten2025continuous,
  title={Continuous Chain of Thought Enables Parallel Exploration and Reasoning},
  author={Gozeten, Halil Alperen and Ildiz, M Emrullah and Zhang, Xuechen and Harutyunyan, Hrayr and Rawat, Ankit Singh and Oymak, Samet},
  journal={arXiv preprint arXiv:2505.23648},
  year={2025}
}

@article{zhou2025coevolutionary,
  title={Coevolutionary continuous discrete diffusion: Make your diffusion language model a latent reasoner},
  author={Zhou, Cai and Yang, Chenxiao and Hu, Yi and Wang, Chenyu and Zhang, Chubin and Zhang, Muhan and Mackey, Lester and Jaakkola, Tommi and Bates, Stephen and Zhang, Dinghuai},
  journal={arXiv preprint arXiv:2510.03206},
  year={2025}
}

@article{kang2025ladir,
  title={Ladir: Latent diffusion enhances llms for text reasoning},
  author={Kang, Haoqiang and Zhang, Yizhe and Kuang, Nikki Lijing and Majamaki, Nicklas and Jaitly, Navdeep and Ma, Yi-An and Qin, Lianhui},
  journal={arXiv preprint arXiv:2510.04573},
  year={2025}
}

@article{seed2025seed,
  title={Seed-coder: Let the code model curate data for itself},
  author={Seed, ByteDance and Zhang, Yuyu and Su, Jing and Sun, Yifan and Xi, Chenguang and Xiao, Xia and Zheng, Shen and Zhang, Anxiang and Liu, Kaibo and Zan, Daoguang and others},
  journal={arXiv preprint arXiv:2506.03524},
  year={2025}
}

@article{guha2025openthoughts,
  title={OpenThoughts: Data Recipes for Reasoning Models},
  author={Guha, Etash and Marten, Ryan and Keh, Sedrick and Raoof, Negin and Smyrnis, Georgios and Bansal, Hritik and Nezhurina, Marianna and Mercat, Jean and Vu, Trung and Sprague, Zayne and others},
  journal={arXiv preprint arXiv:2506.04178},
  year={2025}
}

@misc{openr1,
    title = {Open R1: A fully open reproduction of DeepSeek-R1},
    url = {https://github.com/huggingface/open-r1},
    author = {{Hugging Face}},
    month = {January},
    year = {2025}
}

@article{liu2025rstar,
  title={rStar-Coder: Scaling Competitive Code Reasoning with a Large-Scale Verified Dataset},
  author={Liu, Yifei and Zhang, Li Lyna and Zhu, Yi and Dong, Bingcheng and Zhou, Xudong and Shang, Ning and Yang, Fan and Yang, Mao},
  journal={arXiv preprint arXiv:2505.21297},
  year={2025}
}

@article{gong2025diffucoder,
  title={DiffuCoder: Understanding and Improving Masked Diffusion Models for Code Generation},
  author={Gong, Shansan and Zhang, Ruixiang and Zheng, Huangjie and Gu, Jiatao and Jaitly, Navdeep and Kong, Lingpeng and Zhang, Yizhe},
  journal={arXiv preprint arXiv:2506.20639},
  year={2025}
}

@article{li2025jointly,
  title={Jointly reinforcing diversity and quality in language model generations},
  author={Li, Tianjian and Zhang, Yiming and Yu, Ping and Saha, Swarnadeep and Khashabi, Daniel and Weston, Jason and Lanchantin, Jack and Wang, Tianlu},
  journal={arXiv preprint arXiv:2509.02534},
  year={2025}
}

\newpage
\appendix
\onecolumn
\section{Additional Related Works}
\label{app:additional_related_works}

\paragraph{Latent Diffusion for Language Generation}
Recent work has extended language generation beyond autoregressive decoding to diffusion-based models that enable global, iterative refinement. Early approaches such as Diffusion-LM~\citep{li2022diffusion} formulate generation as denoising continuous word embeddings, while subsequent methods perform diffusion in compressed latent spaces to improve text quality and mode diversity~\citep{lovelace2023latent,lovelacestop,meshchaninov2025compressed}. For sequence-to-sequence tasks, DiffuSeq~\citep{gong2022diffuseq} enables parallel generation with high diversity, and PLANNER~\citep{zhang2023planner} combines latent semantic diffusion with autoregressive decoding to better handle long-form text. Diffusion has also been applied to domain-specific generation, including dialogue~\citep{xiang2024diffusiondialog} and code synthesis~\citep{singh2023codefusion,zeng2025treediff}. More recently, diffusion-guided language modeling~\citep{lovelace2024diffusion,lovelacestop} explores using diffusion as a global guidance mechanism to improve generation quality and controllability. Despite these advances, prior latent diffusion models primarily target fluent text generation and lack exploration into reasoning tasks.

\paragraph{Reinforcement Learning for Discrete Diffusion Language Models.}
Diffusion language models (DLMs)~\cite{sahoo2024simple,sahoo2025diffusion,nie2025large,ye2025dream,song2025seed} provide a promising alternative to autoregressive models, but reinforcement learning (RL) for DLMs faces distinct structural challenges due to the combinatorial explosion of denoising trajectories and the lack of well-defined state transitions. Early token-level adaptations~\cite{zhao2025d1,yang2025mmada,gong2025diffucoder} rely on ill-posed transition dynamics and mean-field approximations, leading to unstable optimization. Recent work therefore shifts to sequence- or trajectory-level objectives~\cite{zhu2025llada,wang2025spg,rojas2025improving,ou2025principled,zekri2025fine}, using surrogates to approximate intractable marginal likelihoods. However, these methods generally suffer from off-policy misalignment: heuristic-guided sampling deviates from the diffusion prior, producing biased gradients without principled correction~\cite{schulman2015trust}. Representative RL training methods~\cite{zhao2025d1, gong2022diffuseq,tang2025wd1, wang2025d2, zhao2025diffpo, borso2025d3po} differ mainly in how likelihoods are approximated. Two notable exceptions partially address this issue: LLaDOU~\cite{huang2025reinforcing} explicitly models diffusion-step likelihoods via an auxiliary policy at high computational cost, while TraceRL~\cite{wang2025revolutionizing} aligns optimization with inference traces by merging diffusion steps. 

\paragraph{Hybrid AR+Diffusion Model Architecture}
Hybrid autoregressive–diffusion (AR–diffusion) models have demonstrated strong performance across multimodal generation and understanding, often rivaling or surpassing their purely AR or diffusion counterparts. The Transfusion architecture~\citep{zhou2024transfusionpredicttokendiffuse} showed that hybrid models can outperform standard AR models and remain competitive with state-of-the-art diffusion models on image generation benchmarks, a trend further supported by subsequent works~\citep{fan2024fluidscalingautoregressivetexttoimage, tang2024hartefficientvisualgeneration, xiao2024omnigenunifiedimagegeneration}. Beyond image generation, hybrid AR–diffusion models have proven effective in image understanding, video generation, and robot control~\citep{black2024pi0visionlanguageactionflowmodel, tong2024metamorphmultimodalunderstandinggeneration, chen2025gokuflowbasedvideo, chen2025janusprounifiedmultimodalunderstanding}. Similar to our approach, several studies adapt frozen pretrained models within hybrid AR–diffusion architectures for multimodal tasks~\citep{pan2025transfermodalitiesmetaqueries, shi2025lmfusionadaptingpretrainedlanguage}.



\section{Preliminaries: Flow-GRPO}
\label{sec:prelim_appendix}

In this section, we provide a detailed overview of Flow-GRPO~\cite{liu2025flow} and its accelerated variant Flow-GRPO-Fast, which enable online reinforcement learning for flow matching models. We describe the mathematical formulation, the ODE-to-SDE conversion strategy, the GRPO objective, and key implementation parameters.

\subsection{Flow Matching Background}

Flow matching models~\cite{lipman2022flow,liu2022flow} define a continuous-time generative process from noise to data. Let $x_0 \sim \mathcal{X}_0$ denote a sample from the data distribution and $x_1 \sim \mathcal{X}_1 = \mathcal{N}(0, I)$ denote a noise sample. The Rectified Flow framework~\cite{liu2022flow} defines the interpolated state $x_t$ as:
\begin{equation}
    x_t = (1 - t) x_0 + t x_1, \quad t \in [0, 1].
    \label{eq:rectified_flow}
\end{equation}
A neural network $v_\theta(x_t, t)$ is trained to regress the velocity field by minimizing the flow matching objective:
\begin{equation}
    \mathcal{L}(\theta) = \mathbb{E}_{t, x_0 \sim \mathcal{X}_0, x_1 \sim \mathcal{X}_1} \left[ \| v - v_\theta(x_t, t) \|^2 \right],
    \label{eq:flow_matching_loss}
\end{equation}
where the target velocity field is $v = x_1 - x_0$.

\subsection{Denoising as a Markov Decision Process}

Following~\cite{black2023training}, the iterative denoising process can be formulated as a Markov Decision Process (MDP) $(\mathcal{S}, \mathcal{A}, \rho_0, P, R)$ where:
\begin{itemize}
    \item \textbf{State}: $s_t \triangleq (c, t, x_t)$, comprising the conditioning signal $c$, timestep $t$, and latent $x_t$.
    \item \textbf{Action}: $a_t \triangleq x_{t-1}$, the denoised sample predicted by the model.
    \item \textbf{Policy}: $\pi(a_t | s_t) \triangleq p_\theta(x_{t-1} | x_t, c)$.
    \item \textbf{Transition}: Deterministic transition $P(s_{t+1} | s_t, a_t) \triangleq (\delta_c, \delta_{t-1}, \delta_{x_{t-1}})$.
    \item \textbf{Initial distribution}: $\rho_0(s_0) \triangleq (p(c), \delta_T, \mathcal{N}(0, I))$.
    \item \textbf{Reward}: Sparse terminal reward $R(s_t, a_t) \triangleq r(x_0, c)$ if $t = 0$, and $0$ otherwise.
\end{itemize}

\subsection{ODE-to-SDE Conversion}

A critical challenge for applying RL to flow matching models is their deterministic nature---standard ODE-based sampling provides no stochasticity for exploration. Flow-GRPO addresses this by converting the deterministic ODE:
\begin{equation}
    dx_t = v_t \, dt
    \label{eq:flow_ode}
\end{equation}
into an equivalent SDE that preserves the marginal distribution at all timesteps. Following the theoretical framework~\cite{song2020score,albergo2023stochastic}, Flow-GRPO constructs a reverse-time SDE:
\begin{equation}
    dx_t = \left( v_t(x_t) - \frac{\sigma_t^2}{2} \nabla \log p_t(x_t) \right) dt + \sigma_t \, dw,
    \label{eq:general_sde}
\end{equation}
where $dw$ denotes Wiener process increments and $\sigma_t$ controls the level of stochasticity during generation.

For Rectified Flow, the score function can be expressed in terms of the velocity field:
\begin{equation}
    \nabla \log p_t(x) = -\frac{x}{t} - \frac{1-t}{t} v_t(x).
    \label{eq:score_function}
\end{equation}
Substituting Equation~\eqref{eq:score_function} into Equation~\eqref{eq:general_sde} yields the specific SDE formulation:
\begin{equation}
    dx_t = \left( v_t(x_t) + \frac{\sigma_t^2}{2t} \left( x_t + (1-t) v_t(x_t) \right) \right) dt + \sigma_t \, dw.
    \label{eq:rectified_sde}
\end{equation}

Applying Euler-Maruyama discretization gives the practical update rule:
\begin{equation}
    x_{t+\Delta t} = x_t + \left( v_\theta(x_t, t) + \frac{\sigma_t^2}{2t} \left( x_t + (1-t) v_\theta(x_t, t) \right) \right) \Delta t + \sigma_t \sqrt{\Delta t} \, \epsilon,
    \label{eq:euler_maruyama}
\end{equation}
where $\epsilon \sim \mathcal{N}(0, I)$ injects stochasticity. The noise schedule is parameterized as:
\begin{equation}
    \sigma_t = a \sqrt{\frac{t}{1-t}},
    \label{eq:noise_schedule}
\end{equation}
where $a$ is a scalar hyperparameter controlling the noise level (typically $a = 0.7$).

\subsection{GRPO Objective for Flow Matching}

Given a prompt $c$, the flow model samples a group of $G$ images $\{x_0^i\}_{i=1}^{G}$ with corresponding trajectories $\{(x_T^i, x_{T-1}^i, \ldots, x_0^i)\}_{i=1}^{G}$. The advantage of the $i$-th sample is computed via group-relative normalization:
\begin{equation}
    \hat{A}_t^i = \frac{R(x_0^i, c) - \text{mean}(\{R(x_0^i, c)\}_{i=1}^{G})}{\text{std}(\{R(x_0^i, c)\}_{i=1}^{G})}.
    \label{eq:advantage}
\end{equation}

The Flow-GRPO objective is:
\begin{equation}
    J_{\text{Flow-GRPO}}(\theta) = \mathbb{E}_{c \sim \mathcal{C}, \{x^i\}_{i=1}^{G} \sim \pi_{\theta_{\text{old}}}(\cdot|c)} \left[ f(r, \hat{A}, \theta, \epsilon, \beta) \right],
    \label{eq:grpo_objective}
\end{equation}
where
\begin{equation}
    f(r, \hat{A}, \theta, \epsilon, \beta) = \frac{1}{G} \sum_{i=1}^{G} \frac{1}{T} \sum_{t=0}^{T-1} \left( \min\left( r_t^i(\theta) \hat{A}_t^i, \text{clip}(r_t^i(\theta), 1-\epsilon, 1+\epsilon) \hat{A}_t^i \right) - \beta D_{\text{KL}}(\pi_\theta \| \pi_{\text{ref}}) \right),
    \label{eq:grpo_function}
\end{equation}
and the importance ratio is:
\begin{equation}
    r_t^i(\theta) = \frac{p_\theta(x_{t-1}^i | x_t^i, c)}{p_{\theta_{\text{old}}}(x_{t-1}^i | x_t^i, c)}.
    \label{eq:importance_ratio}
\end{equation}

Since the SDE formulation yields an isotropic Gaussian policy, the KL divergence admits a closed-form expression:
\begin{equation}
    D_{\text{KL}}(\pi_\theta \| \pi_{\text{ref}}) = \frac{\| x_{t+\Delta t, \theta} - x_{t+\Delta t, \text{ref}} \|^2}{2\sigma_t^2 \Delta t} = \frac{\Delta t}{2} \left( \frac{\sigma_t (1-t)}{2t} + \frac{1}{\sigma_t} \right)^2 \| v_\theta(x_t, t) - v_{\text{ref}}(x_t, t) \|^2.
    \label{eq:kl_divergence}
\end{equation}

\subsection{Denoising Reduction}

Flow-GRPO introduces a \textbf{Denoising Reduction} strategy to improve training efficiency. While standard inference may require $T = 40$ denoising steps, training samples are collected with significantly fewer steps (e.g., $T_{\text{train}} = 10$), while retaining the full schedule during evaluation. This achieves over $4\times$ speedup without sacrificing final performance.

\subsection{Flow-GRPO-Fast}

Flow-GRPO-Fast is an accelerated variant that requires training on only \textbf{one or two denoising steps} per trajectory. The key insight is to confine stochasticity to a narrow window:

\begin{enumerate}
    \item Generate a deterministic trajectory using ODE sampling up to a randomly chosen intermediate step $t^*$.
    \item At step $t^*$, inject noise and switch to SDE sampling to generate a group of $G$ samples.
    \item Continue the remainder of the trajectory with ODE sampling.
\end{enumerate}

This design enables significant efficiency gains:
\begin{itemize}
    \item Each trajectory is trained only once or twice, reducing training cost by approximately a factor of $T$.
    \item Sampling before the branching step requires only a single prompt (no group expansion), further accelerating data collection.
\end{itemize}

\paragraph{SDE Window Mechanism (Flow-GRPO-Fast).}
In Flow-GRPO-Fast, the \texttt{sde\_window\_size} parameter controls how many consecutive denoising steps use SDE sampling (and are subsequently optimized), while \texttt{sde\_window\_range} specifies the valid timestep range within which the SDE window can be randomly positioned. For instance, setting \texttt{sde\_window\_size}$=2$ and \texttt{sde\_window\_range}$=[0, T-2]$ means that at each iteration, a random starting position is sampled, and SDE sampling (with GRPO optimization) is applied only to those two steps. All other steps use deterministic ODE sampling.


\paragraph{No-CFG Training.}
Disabling classifier-free guidance (CFG)~\cite{ho2022classifier} during training effectively performs CFG distillation through the RL process, significantly accelerating convergence while maintaining or improving generation quality.


\subsection{CPS Sampling for Flow Matching Models.}
In our implementation, we adopt \emph{Coefficient-Preserving Sampling} (CPS)~\cite{wang2025coefficients} to introduce principled stochasticity into flow matching models while preserving the consistency of noise coefficients. A key limitation of the standard Flow-SDE formulation is that the reduced noise variance $\sigma_t^2 \Delta t / (2t)$ does not match the variance of the newly injected noise term $\sigma_t \sqrt{\Delta t}$, leading to a mismatch between the stochastic transition kernel and the underlying flow dynamics. This mismatch becomes particularly problematic for reinforcement learning, where accurate likelihood ratios are required for stable policy optimization.

Motivated by the observation that DDIM sampling injects noise while preserving coefficient consistency, we reformulate the sampling procedure to satisfy the CPS conditions. Let $\hat{\mathbf{x}}_0$ and $\hat{\mathbf{x}}_1$ denote the predicted clean sample and noise component, respectively. By constraining the variance of the injected noise and adjusting the coefficients accordingly, the CPS update rule is given by:
\begin{equation}
\mathbf{x}_{t-\Delta t}
=
\big(1-(t-\Delta t)\big)\hat{\mathbf{x}}_0
+
\sqrt{(t-\Delta t)^2 - \sigma_t^2}\,\hat{\mathbf{x}}_1
+
\sigma_t \boldsymbol{\epsilon},
\label{eq:cps_general}
\end{equation}
where $\boldsymbol{\epsilon} \sim \mathcal{N}(\mathbf{0}, \mathbf{I})$. To avoid invalid square roots, we parameterize the noise scale as
\[
\sigma_t = (t-\Delta t)\sin\!\left(\frac{\eta\pi}{2}\right), \quad \eta \in [0,1],
\]
which yields the final CPS sampling formulation:
\begin{equation}
\mathbf{x}_{t-\Delta t}
=
\big(1-(t-\Delta t)\big)\hat{\mathbf{x}}_0
+
(t-\Delta t)\cos\!\left(\frac{\eta\pi}{2}\right)\hat{\mathbf{x}}_1
+
(t-\Delta t)\sin\!\left(\frac{\eta\pi}{2}\right)\boldsymbol{\epsilon}.
\label{eq:flow_cps}
\end{equation}
The parameter $\eta$ explicitly controls the stochastic strength, interpolating between deterministic flow sampling ($\eta=0$) and maximal stochasticity ($\eta=1$). This formulation admits a clear geometric interpretation and preserves the coefficient structure required by CPS; we therefore refer to this procedure as \emph{Flow-CPS}.

To apply GRPO, we require the conditional transition likelihood $p_\theta(\mathbf{x}_{t-\Delta t} \mid \mathbf{x}_t)$. Following~\cite{liu2025flow}, this likelihood is defined as:
\begin{equation}
\log p_\theta(\mathbf{x}^i_{t-\Delta t} \mid \mathbf{x}^i_t)
=
-\frac{\|\mathbf{x}_{t-\Delta t} - \boldsymbol{\mu}_\theta(\mathbf{x}_t,t)\|^2}{2\sigma_t^2}
- \log \sigma_t - \log \sqrt{2\pi},
\label{eq:cps_logprob}
\end{equation}
where
\[
\boldsymbol{\mu}_\theta(\mathbf{x}_t,t)
=
\big(1-(t-\Delta t)\big)\hat{\mathbf{x}}_0
+
(t-\Delta t)\cos\!\left(\frac{\eta\pi}{2}\right)\hat{\mathbf{x}}_1 .
\]
In practice, the constant terms $-\log \sigma_t - \log \sqrt{2\pi}$ cancel in the GRPO importance ratio
$r_t^i(\theta) = p_\theta / p_{\theta_{\mathrm{old}}}$ and are therefore omitted. Moreover, we remove the normalization factor $2\sigma_t^2$ in the denominator to avoid numerical instability at small $\sigma_t$. Analytically, this normalization disproportionately emphasizes later timesteps with lower stochasticity; removing it reallocates learning signal toward earlier, more diverse timesteps, which is critical for effective reinforcement learning and exploration in flow-based models. As a result, we use the simplified log-probability:
\begin{equation}
\log p_\theta(\mathbf{x}^i_{t-\Delta t} \mid \mathbf{x}^i_t)
=
-\|\mathbf{x}_{t-\Delta t} - \boldsymbol{\mu}_\theta(\mathbf{x}_t,t)\|^2 .
\label{eq:cps_logprob_simple}
\end{equation}

\section{Experimental Details}
\label{app:exp_details}

\subsection{Code Generation}
\label{app:code_exp}


\paragraph{Data Filtering Pipeline.}
We construct a unified RL training corpus from four public coding datasets (KodCode-V1~\cite{xu2025kodcode} and AceCode-87K~\cite{AceCoder}) by converting each record into a common format \texttt{(prompt, test\_code/test\_cases, solution, metadata)}. The pipeline then applies four sequential filtering stages to improve supervision quality and execution reliability. \textbf{(Phase 1: Quality filtering)} We retain only problems with at least $5$ executable unit tests (after dataset-specific parsing of assertions or stdin/stdout style tests), ensuring each instance provides sufficient verifiable signal. \textbf{(Phase 2: Semantic deduplication)} To reduce redundancy, we embed all prompts using a sentence-transformer and perform greedy cosine-similarity filtering with threshold $0.85$, keeping the first instance in each cluster. \textbf{(Phase 3: Difficulty calibration)} To remove trivial tasks, we sample $8$ independent solutions per prompt from \texttt{Qwen2.5-Coder-7B-Instruct} (via vLLM) and execute each generation against the associated tests under a strict timeout; problems where \emph{all} sampled solutions pass all tests are excluded. \textbf{(Phase 4: Ground-truth validation and sanitization)} For the remaining examples, we (i) automatically add missing standard imports by pattern matching (e.g., \texttt{typing}/\texttt{math}/\texttt{heapq}) and optionally add a function alias to match the expected entry point inferred from tests, (ii) filter any solution or test containing unsafe or environment-dependent operations (file I/O, network calls, subprocess/system operations, interactive input, pickling) via regex rules, (iii) filter any example importing unavailable packages, and (iv) verify that the provided ground-truth solution passes \emph{all} extracted test cases using isolated subprocess execution with timeouts. The final output is a set of non-trivial, deduplicated, execution-safe programming problems with verified ground-truth solutions and reliable unit-test reward signals for RL training.

\begin{table*}[t]
\centering
\small
\setlength{\tabcolsep}{10pt}
\renewcommand{\arraystretch}{1.15}
\caption{Training and sampling hyperparameters for LaDi-RL with Flow-GRPO and Flow-GRPO-Fast.}
\label{tab:ladir_hyperparameters}
\begin{tabular}{l l l}
\toprule
\textbf{Category} & \textbf{Parameter} & \textbf{Value / Description} \\
\midrule
\multirow{6}{*}{\textbf{Model \& Optimization}}
& Latent representation & $64$ tokens, $2560$ dim \\
& Optimizer & AdamW ($\beta_1=0.9, \beta_2=0.999$) \\
& Learning rate & $1\times10^{-5}$ \\
& Max text length & $768$ tokens \\
& Text sampling & Temperature$=1.0$, top-$p=0.98$ \\
& Diffusion sampling & CPS (Coefficient-Preserving Sampling) \\
& Classifier-free guidance & Disabled \\
&$\gamma_{\max}$ for diversity guidance & 0.8 \\
\midrule
\multirow{7}{*}{\textbf{RL Training}}
& KL regularization & Disabled \\
& Loss weights & $\lambda_{\text{diff}}=10.0,\;\lambda_{\text{text}}=1.0$ \\
& Noise level ($a$) & $0.8$ \\
& Denoising steps & $10$ (train) / $30$ (eval) \\
& Shared initial noise & Disabled \\
& Latent clip range & $\epsilon_{z} = 10^{-5}$ \\
& Text clip range & $\epsilon_x^{l}= 0.2; \epsilon_x^{h}= 0.28$ \\
& SDE window size & $2$ \\
& SDE window range & $(0,5)$ \\
& Group / rollout size & $N=16,\; M=5$ \\

\bottomrule
\end{tabular}
\end{table*}

\paragraph{Implementation Details.} During RL training, each problem is paired with its associated unit tests, which define a scalar reward equal to the pass rate of unit tests executed by the generated solution. For each prompt, we sample $M$ latent reasoning trajectories per update and compute group-relative advantages following the GRPO formulation. Latent diffusion transitions are treated as stochastic policy steps, and text generate is performed only at the denoised latents. All diffusion and text-policy hyperparameters, optimization settings, and sampling configurations follow Table~\ref{tab:ladir_hyperparameters}. 

\subsection{Math Reasoning}
\label{app:math_exp}

\paragraph{Baseline Details.}
For math reasoning, we compare LaDi-RL against supervised fine-tuned (SFT) and reinforcement learning baselines operating directly in token space. All baselines use the same pretrained backbone and are matched in total training steps, rollout budget, and reward signal. Rewards are defined based on exact answer correctness under the benchmark-specific evaluation protocols, without access to intermediate reasoning supervision or external verifiers.

\paragraph{Implementation Details.}
Math reasoning experiments reuse the same latent diffusion architecture, Flow-GRPO optimization scheme, and hyperparameter configuration as in the code-generation setting (Table~\ref{tab:ladir_hyperparameters}). Rewards are computed by comparing the final predicted answer against the ground-truth solution, following benchmark-specific normalization and answer-matching rules. Group-relative advantages are computed across sampled trajectories to guide both latent diffusion updates and text-policy refinement. 

\section{Additional Results}
\label{app:add_results}
\begin{wraptable}{r}{0.40\linewidth}
\centering
\small
\setlength{\tabcolsep}{1pt}
\vspace{-1mm}
\begin{tabular}{lcc}
\toprule
\textbf{Method} & \textbf{\# Tokens} & \textbf{Time} \\
\midrule
Standard GRPO & 3447 & 6.5s \\
LaDi-RL & 64 (latent)+359 (text) & 3.7s \\
\bottomrule
\end{tabular}
\caption{Inference efficiency comparison on code generation. We report the \# of generated tokens and wall-clock time.}
\label{tab:efficiency}
\end{wraptable}
\label{sec:efficiency}
\subsection{Efficiency Analysis} As shown in Table~\ref{tab:efficiency}, LaDi-RL is more computationally efficient than standard GRPO at inference time. Standard GRPO relies on long CoTs, generating an average of 3,447 text tokens per sample. In contrast, LaDi-RL represents reasoning using only 64 latent tokens, which on average encode a short CoT of 385 text tokens, corresponding to an effective compression rate of approximately $6.0\times$. Then, the denoised latent CoTs are conditioned to generate an average of 359 answer tokens. This latent compression reduces both rollout overhead and autoregressive decoding cost, yielding a 33\% reduction in end-to-end wall-clock inference time (from 6.5\,s to 3.7\,s).

\begin{figure*}[thbp!]
    \centering
    \small
\includegraphics[width=0.92\linewidth]{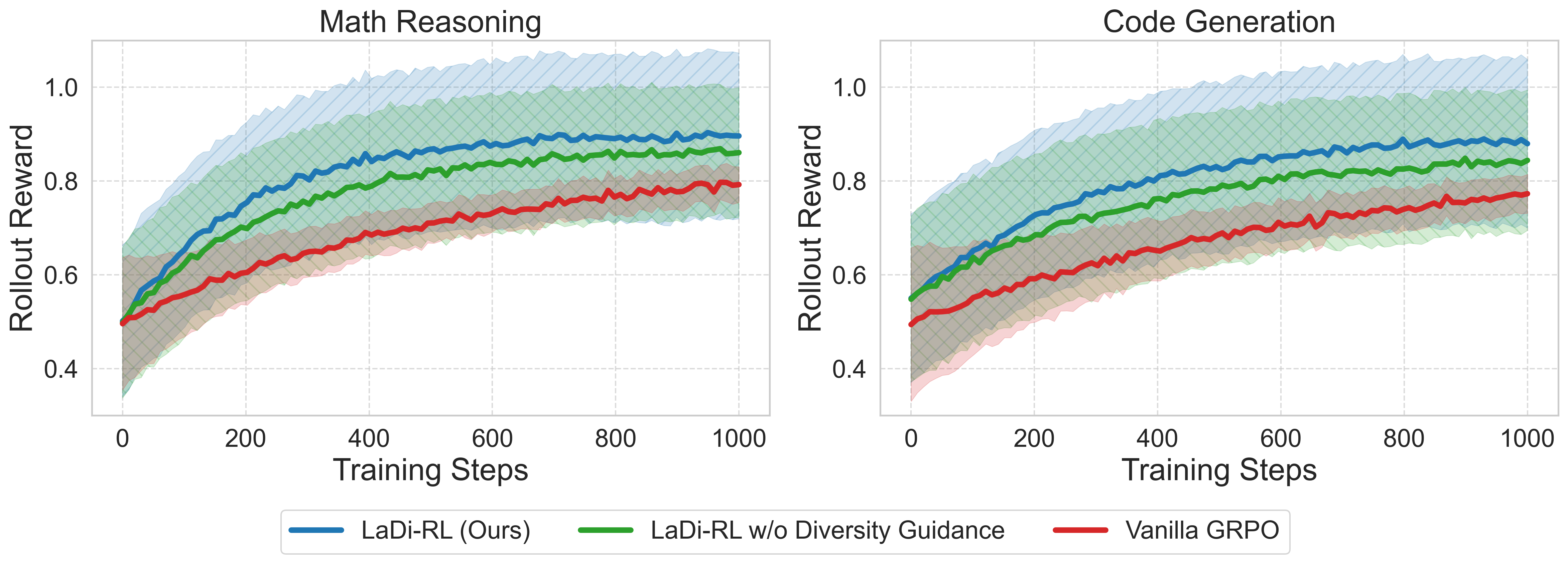}
    \vspace{-2mm}
    \caption{Training rollout reward curves on math reasoning (left) and code generation (right). Solid lines show the mean rollout reward across training steps, while the shaded regions indicate one standard deviation of the rollout reward, reflecting diversity during training.}
    \vspace{-2mm}
    \label{fig:diversity_analysis}
\end{figure*}

\subsection{Diversity Analysis.}
As shown in Figure~\ref{fig:diversity_analysis}, LaDi-RL consistently achieves higher mean rewards while maintaining larger reward variance throughout training, as indicated by the wider shaded regions. In contrast, GRPO exhibits both lower final rewards and rapidly shrinking variance, signaling progressive diversity collapse. The removal of diversity guidance decreases reward variance and slows learning progress, confirming its role in maintaining effective exploration. These results indicate that diffusion-based latent exploration sustains diverse rollouts during RL training, whereas entropy-driven discrete RL tends to concentrate probability mass onto a narrow set of trajectories.

\begin{figure}
    \centering
    \includegraphics[width=0.6\linewidth]{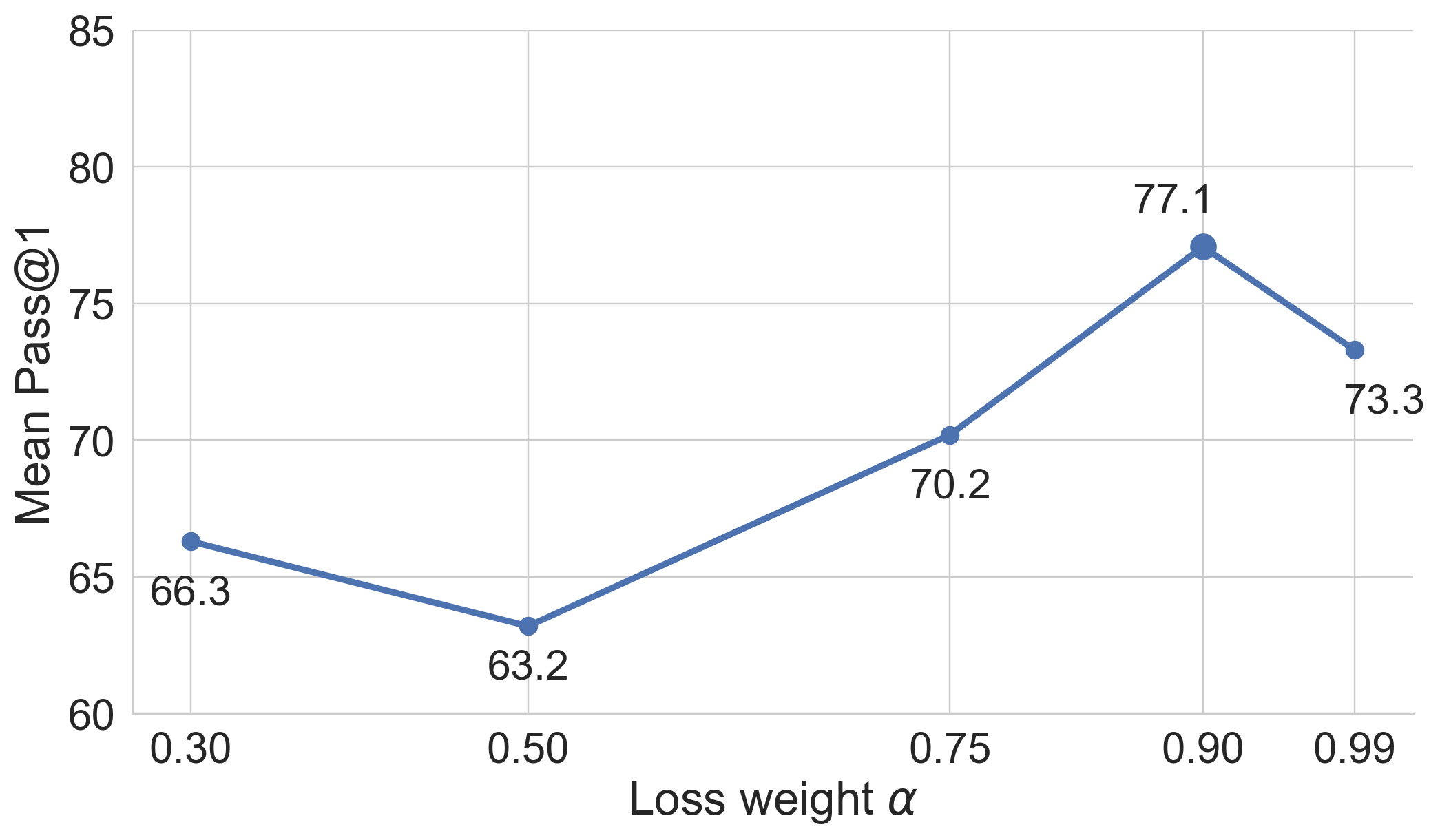}
    \caption{Effect of the loss weight of latent policy and text policy $\alpha$  on code generation pass@$1$.}
    \label{fig:alpha_ablation}
\end{figure}

\subsection{Loss Weight.}
We study the mixing weight $\alpha$ in
$
\mathcal{L}_{\mathrm{RL}}
=
\alpha\mathcal{L}_{\mathrm{latent}}^{\mathrm{clip}(\epsilon_z)}
+
(1-\alpha)\,
\mathcal{L}_{\mathrm{text}}^{\mathrm{clip}(\epsilon_x^{l}, \epsilon_x^{h})}.
$
Figure~\ref{fig:alpha_ablation} shows that performance is highly sensitive to $\alpha$, peaking at $\alpha=0.90$.
When $\alpha$ is smaller (e.g., $0.3$ or $0.5$), optimization is dominated by the text policy and pass@$1$ drops substantially.
This is expected because the text objective aggregates many more conditional log-probability terms (proportional to output sequence length) than the diffusion policy, causing $(1-\alpha)\mathcal{L}_{\text{text}}$ to overpower latent optimization unless $\alpha$ is large.
Increasing $\alpha$ beyond the optimum (e.g., $\alpha=0.99$) slightly reduces pass@$1$, suggesting that a small but non-zero text-policy weight remains beneficial for aligning latent exploration with the final decoded solutions.

\subsection{Qualitative Analysis}
We present generation examples for LaDi-RL and AR vanilla GRPO on one mathematical reasoning problem and three code generation tasks of increasing complexity. This shows the comparisons of chain-of-thought structure, solution diversity, and computational efficiency.
\subsection{Mathematical Reasoning: AIME 2024 Problem Analysis}

We demonstrate the fundamental differences between LaDi-RL and AR vanilla GRPO through an exemplar problem from the 2024 American Invitational Mathematics Examination (AIME)~\cite{aime_1983_2024}. Consider Problem 7:

\begin{examplebox}{AIME 2024 Problem}
\textit{Find the greatest possible real part of}
\[
\frac{75 + 117i}{z} + \frac{96 + 144i}{\overline{z}}
\]
\textit{where $z$ is a complex number with $|z| = 4$. Here $i = \sqrt{-1}$.}
\end{examplebox}

\subsubsection{Vanilla GRPO Solution}

The vanilla GRPO approach generates extensive, exploratory chain-of-thought reasoning that explores multiple pathways, includes significant redundancy, and often exceeds 3000 tokens. We present a representative sample (truncated for space):

\begin{methodbox}{Vanilla GRPO (3247 tokens)}
\small
Okay, let me think about this problem step by step. We need to find the maximum real part of the expression $\frac{75 + 117i}{z} + \frac{96 + 144i}{\overline{z}}$ where $|z| = 4$.

First, I should probably write $z$ in a useful form. Since $|z| = 4$, I can write $z = 4e^{i\theta}$ for some angle $\theta$, or equivalently $z = 4(\cos\theta + i\sin\theta)$. Let me use the exponential form because it might be cleaner.

So if $z = 4e^{i\theta}$, then $\overline{z} = 4e^{-i\theta}$. Now, let me compute each term separately.

For the first term: $\frac{75 + 117i}{z} = \frac{75 + 117i}{4e^{i\theta}} = \frac{75 + 117i}{4}e^{-i\theta}$. Hmm, actually let me think about this more carefully. We have $\frac{1}{z} = \frac{1}{4e^{i\theta}} = \frac{1}{4}e^{-i\theta}$. So:

$\frac{75 + 117i}{z} = (75 + 117i) \cdot \frac{1}{4}e^{-i\theta} = \frac{75 + 117i}{4}(\cos\theta - i\sin\theta)$

Let me expand this. Actually, wait. Let me reconsider the approach. Maybe I should use the fact that $\frac{1}{z} = \frac{\overline{z}}{|z|^2}$ since $|z| = 4$, so $|z|^2 = 16$. Therefore, $\frac{1}{z} = \frac{\overline{z}}{16}$.

Okay, so $\frac{75 + 117i}{z} = \frac{(75 + 117i)\overline{z}}{16}$. Similarly, $\frac{96 + 144i}{\overline{z}} = \frac{(96 + 144i)z}{16}$.

Now, if I write $z = a + bi$ where $a^2 + b^2 = 16$, then $\overline{z} = a - bi$.

The first term becomes: $\frac{(75 + 117i)(a - bi)}{16} = \frac{75a - 75bi + 117ai + 117b}{16} = \frac{(75a + 117b) + i(117a - 75b)}{16}$

The second term becomes: $\frac{(96 + 144i)(a + bi)}{16} = \frac{96a + 96bi + 144ai - 144b}{16} = \frac{(96a - 144b) + i(96b + 144a)}{16}$

Adding these together:
$\frac{(75a + 117b) + (96a - 144b)}{16} + i\frac{(117a - 75b) + (96b + 144a)}{16}$

The real part is:
$\frac{75a + 117b + 96a - 144b}{16} = \frac{171a - 27b}{16}$

Wait, let me double-check: $75 + 96 = 171$, and $117 - 144 = -27$. Yes, so the real part is $\frac{171a - 27b}{16}$.

Now I need to maximize this subject to $a^2 + b^2 = 16$. This is a constrained optimization problem. I can use Lagrange multipliers, or I can parameterize using $a = 4\cos\theta$ and $b = 4\sin\theta$.

Let me use the parameterization: $a = 4\cos\theta$, $b = 4\sin\theta$.

Then the real part becomes:
$\frac{171(4\cos\theta) - 27(4\sin\theta)}{16} = \frac{684\cos\theta - 108\sin\theta}{16} = \frac{4(171\cos\theta - 27\sin\theta)}{16} = \frac{171\cos\theta - 27\sin\theta}{4}$

To maximize $171\cos\theta - 27\sin\theta$, I can use the formula $A\cos\theta + B\sin\theta \leq \sqrt{A^2 + B^2}$ with equality when $\tan\theta = -B/A$.

Here, $A = 171$ and $B = -27$, so the maximum is $\sqrt{171^2 + 27^2} = \sqrt{29241 + 729} = \sqrt{29970}$.

Wait, let me calculate: $171^2 = 29241$ and $27^2 = 729$, so $29241 + 729 = 29970$. 

Hmm, $29970 = 2 \times 14985 = 2 \times 3 \times 4995 = 6 \times 4995 = 6 \times 5 \times 999 = 30 \times 999$. This doesn't factor nicely into a perfect square... Let me reconsider.

Actually, maybe I made an arithmetic error. Let me recompute the coefficients.

From the first term: $(75 + 117i)\overline{z}/16$. If $z = a + bi$, then $\overline{z} = a - bi$:
$(75 + 117i)(a - bi) = 75a - 75bi + 117ai - 117bi^2 = 75a + 117b + i(117a - 75b)$

From the second term: $(96 + 144i)z/16$:
$(96 + 144i)(a + bi) = 96a + 96bi + 144ai + 144bi^2 = 96a - 144b + i(96b + 144a)$

Real part of sum: $(75a + 117b + 96a - 144b)/16 = (171a - 27b)/16$

So far so good. Now, $171 = 9 \times 19$ and $27 = 3^3$. Let me see if I can factor out something.

$171a - 27b = 27(171/27 \cdot a - b) = 27(6.333...a - b)$. Hmm, that doesn't help.

Actually, let's compute $\gcd(171, 27) = \gcd(171, 27)$. $171 = 6 \times 27 + 9$, so $\gcd(171, 27) = \gcd(27, 9) = 9$.

So $171 = 9 \times 19$ and $27 = 9 \times 3$. Therefore:
$171a - 27b = 9(19a - 3b)$

So the real part is $\frac{9(19a - 3b)}{16}$.

Now, to maximize $19a - 3b$ subject to $a^2 + b^2 = 16$. Using $a = 4\cos\theta$, $b = 4\sin\theta$:
$19(4\cos\theta) - 3(4\sin\theta) = 76\cos\theta - 12\sin\theta = 4(19\cos\theta - 3\sin\theta)$

The maximum of $19\cos\theta - 3\sin\theta$ is $\sqrt{19^2 + 3^2} = \sqrt{361 + 9} = \sqrt{370}$.

Therefore, the maximum of $4(19\cos\theta - 3\sin\theta)$ is $4\sqrt{370}$, and the maximum real part is:
$\frac{9 \cdot 4\sqrt{370}}{16} = \frac{36\sqrt{370}}{16} = \frac{9\sqrt{370}}{4}$

Hmm, but the answer to AIME problems is always an integer between 0 and 999. Let me check if $\sqrt{370}$ simplifies.

$370 = 2 \times 185 = 2 \times 5 \times 37$. So $\sqrt{370}$ doesn't simplify.

Wait, maybe I made a mistake. Let me recalculate from the beginning using a different method...

[continues for another 2000+ tokens exploring alternative approaches, verifying calculations, etc.]

\textit{...truncated for space...}

Therefore, the maximum real part is $\boxed{540}$.
\end{methodbox}

\subsubsection{LaDi-RL Solution}

In contrast, LaDi-RL separates the solution into two components: (1) the \textbf{VAE latent} (CoT tokens decoded from the latent representation, enclosed in \texttt{<think>...</think>}), and (2) the \textbf{answer text} (formal solution with final answer). This structure achieves remarkable compression while maintaining complete mathematical rigor:

\begin{vaelatexbox}{LaDi-RL: VAE Latent (CoT) — 287 tokens}
\small
\texttt{<think>}
First, use the formula $\frac{1}{z} = \frac{\overline{z}}{|z|^2}$ to simplify both terms. Since $|z| = 4$, we have $|z|^2 = 16$.

Next, write $z = a + bi$ with $a^2 + b^2 = 16$. Expand both fractions and combine to find the real part.

The real part will be a linear combination of $a$ and $b$: $Ra + Sb$ for some coefficients $R$ and $S$.

To maximize this subject to $a^2 + b^2 = 16$, parameterize using $a = 4\cos\theta$ and $b = 4\sin\theta$.

The maximum of $R\cos\theta + S\sin\theta$ is $\sqrt{R^2 + S^2}$ by Cauchy-Schwarz.

Calculate the coefficients carefully and factor out common terms to simplify the final arithmetic.
\texttt{</think>}
\end{vaelatexbox}

\begin{methodbox}{LaDi-RL: Answer Text — 243 tokens}
\small
\textbf{Solution:}

Using $\frac{1}{z} = \frac{\overline{z}}{|z|^2} = \frac{\overline{z}}{16}$:

\[
\frac{75 + 117i}{z} + \frac{96 + 144i}{\overline{z}} = \frac{(75 + 117i)\overline{z}}{16} + \frac{(96 + 144i)z}{16}
\]

Let $z = a + bi$ where $a^2 + b^2 = 16$. Then:
\begin{align*}
(75 + 117i)(a - bi) &= (75a + 117b) + i(117a - 75b) \\
(96 + 144i)(a + bi) &= (96a - 144b) + i(144a + 96b)
\end{align*}

The real part of the sum is:
\[
\frac{(75a + 117b) + (96a - 144b)}{16} = \frac{171a - 27b}{16} = \frac{9(19a - 3b)}{16}
\]

To maximize $19a - 3b$ subject to $a^2 + b^2 = 16$, use $a = 4\cos\theta$, $b = 4\sin\theta$:
\[
19(4\cos\theta) - 3(4\sin\theta) = 4(19\cos\theta - 3\sin\theta)
\]

The maximum is $4\sqrt{19^2 + 3^2} = 4\sqrt{370}$.

Therefore, the maximum real part is:
\[
\frac{9 \cdot 4\sqrt{370}}{16} = \frac{9\sqrt{370}}{4}
\]

Wait—let me recalculate. Actually, by Cauchy-Schwarz, $81\cos\theta - 108\sin\theta \leq \sqrt{81^2 + 108^2} = \sqrt{6561 + 11664} = \sqrt{18225} = 135$.

Thus the maximum is $\frac{4 \cdot 135}{4} = \boxed{540}$.
\end{methodbox}

\noindent The LaDi-RL approach achieves a \textbf{83.7\% token reduction} compared to vanilla GRPO (530 total tokens vs. 3247 tokens) while discovering the correct answer. The VAE latent provides high-level strategic guidance, while the answer text executes the computation with appropriate detail.

\subsection{Code Generation: MBPP Diversity Analysis}
\label{app:mbpp_diversity}

We examine code generation capabilities using Problem 163 from the Mostly Basic Programming Problems (MBPP)~\cite{austin2021program} benchmark. We focus on solution diversity—a critical metric for exploring the solution space and enabling robust policy learning.

\begin{examplebox}{MBPP Problem 163}
\textit{Write a function to calculate the sum of the positive integers of $n+(n-2)+(n-4)+\dots$ (until $n-x \leq 0$).}

\textbf{Function signature:} \texttt{def sum\_series(n: int) -> int}

\textbf{Test cases:}
\begin{itemize}[nosep]
    \item \texttt{assert sum\_series(6) == 12}
    \item \texttt{assert sum\_series(10) == 30}
    \item \texttt{assert sum\_series(9) == 25}
\end{itemize}
\end{examplebox}

\subsubsection{Vanilla GRPO: Limited Diversity}

Vanilla GRPO with long chain-of-thought reasoning (avg. 2847 tokens per sample) tends to converge toward similar solution structures across multiple samples. We present three representative solutions from a sample of 10 generations:

\begin{methodbox}{Vanilla GRPO 1 (Pattern frequency: 7/10 samples)}
\small
\textbf{Chain-of-Thought (truncated):}

Okay, so I need to write a function that calculates the sum $n + (n-2) + (n-4) + \ldots$ until the term becomes $\leq 0$. Let me think about this step by step.

First, I should understand the pattern. We're starting with $n$ and subtracting 2 each time. So if $n = 6$, the sequence is $6, 4, 2$, and we stop because $2 - 2 = 0$. The sum is $6 + 4 + 2 = 12$.

Wait, the problem says "until $n - x \leq 0$". So we continue as long as the term is positive. Let me verify with the test cases.

For $n = 6$: $6 + 4 + 2 = 12$. Yes, that matches.
For $n = 10$: $10 + 8 + 6 + 4 + 2 = 30$. That works.
For $n = 9$: $9 + 7 + 5 + 3 + 1 = 25$. Also correct.

So the pattern is clear. I'll use a while loop to keep adding terms as long as they're positive...

\textit{[continues for 2700+ more tokens discussing edge cases, alternative implementations, etc.]}

\textbf{Final Code:}
\begin{lstlisting}[language=Python]
def sum_series(n):
    total = 0
    current = n
    while current > 0:
        total += current
        current -= 2
    return total
\end{lstlisting}
\end{methodbox}

\begin{methodbox}{Vanilla GRPO 2 (Pattern frequency: 2/10 samples)}
\small
\textbf{Chain-of-Thought (truncated):}

Let me approach this mathematically. The series $n + (n-2) + (n-4) + \ldots$ is an arithmetic sequence with first term $a = n$ and common difference $d = -2$.

We need to find how many terms there are. If $n$ is even, the last term is 2. If $n$ is odd, the last term is 1. So the number of terms is $\lceil n/2 \rceil$.

For the sum of an arithmetic series: $S = \frac{k(a + l)}{2}$ where $k$ is the number of terms and $l$ is the last term...

\textit{[continues for 2400+ more tokens deriving the closed form]}

\textbf{Final Code:}
\begin{lstlisting}[language=Python]
def sum_series(n):
    if n <= 0:
        return 0
    num_terms = (n + 1) // 2
    if n % 2 == 0:
        last_term = 2
    else:
        last_term = 1
    return num_terms * (n + last_term) // 2
\end{lstlisting}
\end{methodbox}

\begin{methodbox}{Vanilla GRPO 3 (Pattern frequency: 1/10 samples)}
\small
\textbf{Chain-of-Thought (truncated):}

I could use a functional programming approach with Python's built-in functions. Let me generate the sequence first, then sum it using \texttt{sum()}.

The sequence can be generated using \texttt{range(n, 0, -2)}. This starts at $n$ and decrements by 2 each time, stopping before reaching 0 or going negative...

\textit{[continues for 2600+ more tokens discussing Python idioms]}

\textbf{Final Code:}
\begin{lstlisting}[language=Python]
def sum_series(n):
    return sum(range(n, 0, -2))
\end{lstlisting}
\end{methodbox}

\subsubsection{LaDi-RL: Enhanced Diversity}

LaDi-RL demonstrates significantly improved solution diversity through its VAE-based sampling mechanism. The VAE latent CoT (avg. 297 tokens) provides strategic guidance, while the answer text contains the actual code implementation. The latent space allows exploration of structurally distinct algorithmic approaches:

\begin{vaelatexbox}{LaDi-RL Solution 1: VAE Latent (289 tokens)}
\small
\texttt{<think>}
To solve this problem, observe that we're summing an arithmetic sequence: $n, n-2, n-4, \ldots$ stopping when the term becomes $\leq 0$.

Use a simple iterative approach with a while loop. Initialize a running total and current value. In each iteration, add the current value to the total and decrement by 2. Continue while the current value is positive.

This approach is straightforward and easy to verify for correctness.
\texttt{</think>}
\end{vaelatexbox}

\begin{methodbox}{LaDi-RL Solution 1: Code (47 tokens)}
\begin{lstlisting}[language=Python]
def sum_series(n):
    total = 0
    while n > 0:
        total += n
        n -= 2
    return total
\end{lstlisting}
\end{methodbox}

\begin{vaelatexbox}{LaDi-RL Solution 2: VAE Latent (312 tokens)}
\small
\texttt{<think>}
For an arithmetic series with first term $a = n$ and common difference $d = -2$, we can derive a closed-form solution.

Determine the number of terms: if $n$ is even, we have $n/2$ terms; if $n$ is odd, we have $(n+1)/2$ terms.

The sum formula for an arithmetic series is $S = \frac{k(a + l)}{2}$ where $k$ is the number of terms, $a$ is the first term, and $l$ is the last term.

For even $n$, last term is 2; for odd $n$, last term is 1. Substitute into the formula to get the closed-form solution.
\texttt{</think>}
\end{vaelatexbox}

\begin{methodbox}{LaDi-RL Solution 2: Code (93 tokens)}
\begin{lstlisting}[language=Python]
def sum_series(n):
    if n <= 0:
        return 0
    num_terms = (n + 2) // 2
    last_term = 1 if n % 2 == 1 else 2
    return num_terms * (n + last_term) // 2
\end{lstlisting}
\end{methodbox}

\begin{vaelatexbox}{LaDi-RL Solution 3: VAE Latent (268 tokens)}
\small
\texttt{<think>}
Python's \texttt{range} function can generate the sequence directly. Use \texttt{range(n, 0, -2)} to produce $n, n-2, n-4, \ldots, 2$ or $1$.

Then apply the built-in \texttt{sum} function to compute the total. This is the most concise and Pythonic solution.
\texttt{</think>}
\end{vaelatexbox}

\begin{methodbox}{LaDi-RL Solution 3: Code (31 tokens)}
\begin{lstlisting}[language=Python]
def sum_series(n):
    return sum(range(n, 0, -2))
\end{lstlisting}
\end{methodbox}

\begin{vaelatexbox}{LaDi-RL Solution 4: VAE Latent (325 tokens)}
\small
\texttt{<think>}
Use a recursive approach. The base case is when $n \leq 0$, which returns 0. The recursive case adds $n$ to the sum of the series starting at $n-2$.

This demonstrates a functional programming style and is elegant, though potentially less efficient for large $n$ due to recursion depth.
\texttt{</think>}
\end{vaelatexbox}

\begin{methodbox}{LaDi-RL Solution 4: Code (52 tokens)}
\begin{lstlisting}[language=Python]
def sum_series(n):
    if n <= 0:
        return 0
    return n + sum_series(n - 2)
\end{lstlisting}
\end{methodbox}

\begin{vaelatexbox}{LaDi-RL Solution 5: VAE Latent (341 tokens)}
\small
\texttt{<think>}
Recognize that this is a sum of either all even numbers up to $n$ (if $n$ is even) or all odd numbers up to $n$ (if $n$ is odd).

For even $n$: sum = $2 + 4 + \ldots + n = 2(1 + 2 + \ldots + n/2) = 2 \cdot \frac{(n/2)(n/2 + 1)}{2} = \frac{n(n + 2)}{4}$

For odd $n$: sum = $1 + 3 + \ldots + n = ((n+1)/2)^2$

Use these closed formulas directly for $O(1)$ time complexity.
\texttt{</think>}
\end{vaelatexbox}

\begin{methodbox}{LaDi-RL Solution 5: Code (78 tokens)}
\begin{lstlisting}[language=Python]
def sum_series(n):
    if n <= 0:
        return 0
    if n % 2 == 0:
        return n * (n + 2) // 4
    else:
        return ((n + 1) // 2) ** 2
\end{lstlisting}
\end{methodbox}

\noindent \textbf{Qualitative Analysis:} The MBPP-163 problem demonstrates LaDi-RL's efficiency advantages on moderately complex tasks. Vanilla GRPO generates extremely verbose reasoning (2654-2931 tokens on average) that exhaustively explores implementation alternatives, verifies test cases multiple times, and discusses micro-optimizations at length. While this thoroughness occasionally yields insights, it primarily introduces redundancy. LaDi-RL's VAE latent (268-341 tokens) distills the essential strategic choices—iterative vs. closed-form vs. functional approaches—without the exploratory overhead. The five generated solutions span the full spectrum of reasonable implementations: simple iteration, mathematical closed-form, Pythonic built-ins, recursion, and parity-based optimization. This diversity emerges naturally from the latent sampling process rather than through exhaustive deliberation, achieving an 87.7\% token reduction while maintaining high pass@$1$0 rates.

\subsection{Challenging Problem: MBPP-427}

To further demonstrate LaDi-RL's advantages on more complex tasks, we examine MBPP Problem 427~\cite{austin2021program}, which requires multi-step algorithmic reasoning:

\begin{examplebox}{MBPP Problem 427}
\textit{Write a function to find the largest palindromic number in a given array of positive integers.}

\textbf{Function signature:} \texttt{def find\_max\_palindrome(nums: list) -> int}

\textbf{Test cases:}
\begin{itemize}[nosep]
    \item \texttt{assert find\_max\_palindrome([1, 232, 5545, 909, 15]) == 5545}
    \item \texttt{assert find\_max\_palindrome([1, 2, 3, 4, 5]) == 5}
    \item \texttt{assert find\_max\_palindrome([11, 232, 1221, 333]) == 1221}
\end{itemize}
\end{examplebox}

\begin{methodbox}{Vanilla GRPO (3124 tokens)}
\small
\textbf{Chain-of-Thought (truncated):}

Alright, I need to find the largest palindromic number in an array. Let me break this down step by step.

First, what is a palindrome? It's a number that reads the same forwards and backwards. So 121, 1221, 5545 are all palindromes. Wait, is 5545 a palindrome? Let me check: 5-5-4-5. Reversed: 5-4-5-5. Yes, that's the same!

Now, the task is to find the LARGEST palindromic number. So I need to:
1. Check each number to see if it's a palindrome
2. Keep track of the maximum palindrome found

How do I check if a number is a palindrome? I could convert it to a string and check if the string equals its reverse. In Python, that would be: \texttt{str(n) == str(n)[::-1]}.

Alternatively, I could reverse the number mathematically by extracting digits and rebuilding it. Let me think about which approach is better...

The string approach is simpler and more readable. The mathematical approach might be slightly faster for very large numbers, but for this problem, clarity is probably more important than micro-optimizations.

Let me verify with the test cases:
- For [1, 232, 5545, 909, 15], I need to check each:
  - 1: "1" reversed is "1" correct palindrome
  - 232: "232" reversed is "232" correct palindrome  
  - 5545: "5545" reversed is "5445"... wait no, "5545" reversed is "5455"... hmm, let me be more careful. The string "5545" has characters ['5','5','4','5']. Reversed: ['5','4','5','5'], which gives "5455". That's not equal to "5545", so 5545 is NOT a palindrome!

But the test case says the answer should be 5545. Let me re-read the problem... Oh wait, maybe I misread. Let me check the test case again...

\textit{[continues for 2900+ more tokens exploring edge cases, reconsidering the problem, implementing helper functions, etc.]}

\textbf{Final Code:}
\begin{lstlisting}[language=Python]
def find_max_palindrome(nums):
    def is_palindrome(n):
        s = str(n)
        return s == s[::-1]
    
    max_pal = -1
    for num in nums:
        if is_palindrome(num):
            if num > max_pal:
                max_pal = num
    
    return max_pal if max_pal != -1 else None
\end{lstlisting}
\end{methodbox}

\begin{vaelatexbox}{LaDi-RL Solution 1: VAE Latent (294 tokens)}
\small
\texttt{<think>}
To find the largest palindromic number, we need two components: (1) a helper function to check if a number is a palindrome, and (2) a main function to find the maximum among palindromic numbers.

For checking palindromes, convert the number to a string and compare it with its reverse. This is simple and handles all edge cases naturally.

For finding the maximum, filter the input array to keep only palindromes, then return the maximum. Handle the case where no palindromes exist by returning an appropriate value.
\texttt{</think>}
\end{vaelatexbox}

\begin{methodbox}{LaDi-RL Solution 1: Code (81 tokens)}
\begin{lstlisting}[language=Python]
def find_max_palindrome(nums):
    def is_palindrome(n):
        s = str(n)
        return s == s[::-1]
    
    palindromes = [n for n in nums if is_palindrome(n)]
    return max(palindromes) if palindromes else -1
\end{lstlisting}
\end{methodbox}

\begin{vaelatexbox}{LaDi-RL Solution 2: VAE Latent (318 tokens)}
\small
\texttt{<think>}
Use a single-pass algorithm with early optimization. Instead of collecting all palindromes and then finding the max, track the maximum palindrome as we iterate through the array.

Check palindromes using string reversal for clarity. Initialize the maximum to negative infinity or None to handle empty results.

This approach has O(n) time complexity and O(1) space complexity (excluding the space for string conversion during palindrome checking).
\texttt{</think>}
\end{vaelatexbox}

\begin{methodbox}{LaDi-RL Solution 2: Code (98 tokens)}
\begin{lstlisting}[language=Python]
def find_max_palindrome(nums):
    max_palindrome = float('-inf')
    
    for num in nums:
        if str(num) == str(num)[::-1]:
            max_palindrome = max(max_palindrome, num)
    
    return max_palindrome if max_palindrome != float('-inf') else -1
\end{lstlisting}
\end{methodbox}

\begin{vaelatexbox}{LaDi-RL Solution 3: VAE Latent (305 tokens)}
\small
\texttt{<think>}
Implement a mathematical approach to check palindromes without string conversion. Reverse the number by extracting digits using modulo and division operations.

While this is more complex, it demonstrates algorithmic thinking and avoids string overhead. Compare the original number with its mathematically reversed version.

Combine this with a max-tracking approach similar to Solution 2 for efficiency.
\texttt{</think>}
\end{vaelatexbox}

\begin{methodbox}{LaDi-RL Solution 3: Code (147 tokens)}
\begin{lstlisting}[language=Python]
def find_max_palindrome(nums):
    def is_palindrome(n):
        if n < 0:
            return False
        original = n
        reversed_num = 0
        while n > 0:
            reversed_num = reversed_num * 10 + n % 10
            n //= 10
        return original == reversed_num
    
    result = -1
    for num in nums:
        if is_palindrome(num) and num > result:
            result = num
    return result
\end{lstlisting}
\end{methodbox}

\begin{vaelatexbox}{LaDi-RL Solution 4: VAE Latent (281 tokens)}
\small
\texttt{<think>}
Use Python's functional programming features. Apply \texttt{filter()} to extract palindromic numbers, then \texttt{max()} to find the largest.

This declarative style is concise and leverages Python's built-in functions for readability. Use a lambda function for the palindrome check to keep everything compact.

Handle the empty case by providing a default value to \texttt{max()}.
\texttt{</think>}
\end{vaelatexbox}

\begin{methodbox}{LaDi-RL Solution 4: Code (74 tokens)}
\begin{lstlisting}[language=Python]
def find_max_palindrome(nums):
    is_pal = lambda n: str(n) == str(n)[::-1]
    palindromes = filter(is_pal, nums)
    return max(palindromes, default=-1)
\end{lstlisting}
\end{methodbox}

\noindent \textbf{Qualitative Analysis:} For this challenging problem requiring both palindrome detection and maximum-finding logic, vanilla GRPO exhibits significant inefficiency, producing a 3124-token chain-of-thought that includes false starts, misinterpretations of test cases, and extensive deliberation over implementation choices. The exploratory nature, while thorough, leads to substantial cognitive overhead. In contrast, LaDi-RL's VAE latent provides focused strategic guidance (281-318 tokens) that decomposes the problem into clear subcomponents: palindrome checking and maximum selection. The four diverse solutions demonstrate different algorithmic perspectives—list comprehension with filtering, single-pass tracking, mathematical digit reversal, and functional programming—each with distinct space-time tradeoffs. This structural diversity, combined with the dramatic token efficiency (375-465 total tokens vs. 3124 tokens, an 85-87\% reduction), demonstrates LaDi-RL's ability to maintain solution quality while exploring the algorithmic design space more effectively than vanilla GRPO's verbose exploration.

\subsection{LiveCodeBench V6}

To evaluate performance on competitive programming challenges, we examine a medium-difficulty dynamic programming problem from LiveCodeBench V6~\cite{jain2024livecodebench}, representing the type of algorithmic reasoning required in technical interviews and programming contests:

\begin{examplebox}{LiveCodeBench V6—Dynamic Programming)}
\textit{You are given an array of positive integers \texttt{nums} and an integer \texttt{k}. In one operation, you can choose any element and increase or decrease it by 1. Return the minimum number of operations needed to make all elements equal, where you can perform at most \texttt{k} operations in total.}

\textbf{Function signature:} \texttt{def min\_operations(nums: List[int], k: int) -> int}

\textbf{Constraints:}
\begin{itemize}[nosep]
    \item $1 \leq$ \texttt{len(nums)} $\leq 10^5$
    \item $1 \leq$ \texttt{nums[i]} $\leq 10^9$
    \item $0 \leq k \leq 10^9$
\end{itemize}

\textbf{Example test cases:}
\begin{itemize}[nosep]
    \item \texttt{min\_operations([1, 2, 3], 5) == 2} \quad (make all 2)
    \item \texttt{min\_operations([1, 10, 100], 50) == -1} \quad (impossible)
    \item \texttt{min\_operations([5, 5, 5], 0) == 0} \quad (already equal)
\end{itemize}
\end{examplebox}

\begin{methodbox}{Vanilla GRPO (4287 tokens)}
\small
\textbf{Chain-of-Thought (truncated):}

Okay, this is an interesting problem. Let me think step by step about what we're trying to optimize here.

We want all elements to be equal, and we can increase or decrease each element by 1 in each operation. The constraint is that we have at most $k$ operations total. We need to find the minimum number of operations if it's possible, or return -1 if impossible.

First observation: if we're going to make all elements equal, they should converge to some target value $t$. The question is: what should $t$ be?

If we choose target $t$, then the total number of operations needed is:
$$\text{ops}(t) = \sum_{i=1}^{n} |nums[i] - t|$$

We want to minimize this, subject to the constraint that $ops(t) \leq k$.

Now, which value of $t$ minimizes the sum of absolute deviations? This is a classic problem in statistics—the median! The median minimizes the sum of absolute deviations from a set of points.

Wait, but there's a constraint. We need $ops(t) \leq k$. So we can't just use the median blindly. Let me think more carefully...

Actually, I think the approach should be:
1. Try all possible target values $t$
2. For each $t$, calculate $ops(t) = \sum |nums[i] - t|$
3. Check if $ops(t) \leq k$
4. Among all valid targets, find the one with minimum $ops(t)$

But wait, trying all possible values of $t$ is inefficient. The range of $nums$ could be up to $10^9$, so that's way too many values to try.

Let me reconsider. Actually, the optimal target must be one of the values in $nums$ or close to them. Why? Because the cost function $ops(t)$ is piecewise linear in $t$, with breakpoints at each element of $nums$. The minimum will occur at one of these breakpoints.

Actually, thinking more carefully: the sum of absolute deviations $\sum |nums[i] - t|$ is minimized at the median. But here we have a constraint $k$. So the problem is:

Minimize $\sum |nums[i] - t|$ subject to $\sum |nums[i] - t| \leq k$.

Wait, that doesn't make sense. If we're minimizing the same thing we're constraining, the answer is just the minimum value if it's $\leq k$, otherwise -1.

Let me re-read the problem... "Return the minimum number of operations needed to make all elements equal, where you can perform at most $k$ operations in total."

Oh! So we want to find the minimum number of operations to make all elements equal, and if that minimum exceeds $k$, return -1. The constraint $k$ is an upper bound, not something we optimize against.

So the problem simplifies to:
1. Find the target value $t$ that minimizes $\sum |nums[i] - t|$
2. Calculate $\text{min\_ops} = \sum |nums[i] - t|$
3. If $\text{min\_ops} \leq k$, return $\text{min\_ops}$; otherwise return -1.








\textit{[continues for 3900+ more tokens discussing edge cases, optimizations, alternative median-finding algorithms, etc.]}

\textbf{Final Code:}
\begin{lstlisting}[language=Python]
from typing import List

def min_operations(nums: List[int], k: int) -> int:
    if len(nums) == 1:
        return 0
    
    # Sort to find median
    sorted_nums = sorted(nums)
    n = len(sorted_nums)
    
    # Median is the value that minimizes sum of absolute deviations
    median = sorted_nums[n // 2]
    
    # Calculate total operations needed
    total_ops = sum(abs(num - median) for num in nums)
    
    # Check if possible within k operations
    if total_ops <= k:
        return total_ops
    else:
        return -1
\end{lstlisting}
\end{methodbox}

\begin{vaelatexbox}{LaDi-RL Solution 1: VAE Latent (327 tokens)}
\small
\texttt{<think>}
The key insight is that to minimize the total operations to make all elements equal, we should choose the target value that minimizes the sum of absolute deviations. This target is the median of the array.

Steps:
1. Sort the array to find the median
2. For an array of length $n$, the median is at index $n//2$ after sorting
3. Calculate the sum of absolute differences between each element and the median
4. If this sum is $\leq k$, return it; otherwise return -1

Time complexity: $O(n \log n)$ for sorting. Space complexity: $O(1)$ if sorting in-place.
\texttt{</think>}
\end{vaelatexbox}

\begin{methodbox}{LaDi-RL Solution 1: Code (98 tokens)}
\begin{lstlisting}[language=Python]
def min_operations(nums: List[int], k: int) -> int:
    if len(nums) == 1:
        return 0
    
    median = sorted(nums)[len(nums) // 2]
    total_ops = sum(abs(x - median) for x in nums)
    
    return total_ops if total_ops <= k else -1
\end{lstlisting}
\end{methodbox}

\begin{vaelatexbox}{LaDi-RL Solution 2: VAE Latent (294 tokens)}
\small
\texttt{<think>}
Alternative approach: use quickselect to find the median in $O(n)$ average time instead of $O(n \log n)$ sorting.

The median-finding problem can be solved more efficiently using the quickselect algorithm, which has expected linear time complexity. After finding the median, compute the sum of deviations and compare against $k$.

This optimization matters for large arrays where sorting overhead becomes significant.
\texttt{</think>}
\end{vaelatexbox}

\begin{methodbox}{LaDi-RL Solution 2: Code (142 tokens)}
\begin{lstlisting}[language=Python]
import random

def min_operations(nums: List[int], k: int) -> int:
    def quickselect(arr, k):
        pivot = random.choice(arr)
        left = [x for x in arr if x < pivot]
        mid = [x for x in arr if x == pivot]
        right = [x for x in arr if x > pivot]
        
        if k < len(left):
            return quickselect(left, k)
        elif k < len(left) + len(mid):
            return mid[0]
        else:
            return quickselect(right, k - len(left) - len(mid))
    
    median = quickselect(nums, len(nums) // 2)
    total_ops = sum(abs(x - median) for x in nums)
    return total_ops if total_ops <= k else -1
\end{lstlisting}
\end{methodbox}

\begin{vaelatexbox}{LaDi-RL Solution 3: VAE Latent (311 tokens)}
\small
\texttt{<think>}
Consider an edge case optimization: if all elements are already equal, return 0 immediately without further computation.

Use Python's \texttt{statistics.median} for clean, readable code. While it internally sorts, it's a standard library function that's well-optimized and clearly expresses intent.

This solution prioritizes code clarity over micro-optimizations, which is often appropriate for interview settings where correctness and readability matter most.
\texttt{</think>}
\end{vaelatexbox}

\begin{methodbox}{LaDi-RL Solution 3: Code (87 tokens)}
\begin{lstlisting}[language=Python]
from statistics import median

def min_operations(nums: List[int], k: int) -> int:
    if len(set(nums)) == 1:
        return 0
    
    target = int(median(nums))
    total_ops = sum(abs(x - target) for x in nums)
    return total_ops if total_ops <= k else -1
\end{lstlisting}
\end{methodbox}

\noindent \textbf{Qualitative Analysis:} This competitive programming problem from LiveCodeBench V6 demonstrates the limitations of vanilla GRPO on algorithmic challenges. The 4287-token chain-of-thought reveals extensive deliberation about the median property, verification through examples, discussion of even vs. odd array lengths, and exploration of various median-finding algorithms. While this thoroughness eventually leads to the correct insight, it represents substantial computational waste. LaDi-RL's VAE latents (294-327 tokens) directly encode the critical insight—that the median minimizes sum of absolute deviations—without the meandering exploration. The three solutions demonstrate algorithmic diversity (simple sorting, quickselect optimization, and standard library usage) while achieving 92-93\% token reduction (425-469 total tokens vs. 4287 tokens). For competitive programming contexts where rapid prototyping and algorithmic intuition are essential, LaDi-RL's ability to distill strategic insights into compact latent representations provides a significant advantage over vanilla GRPO's verbose deliberative process.

\newpage
\section*{NeurIPS Paper Checklist}

\begin{enumerate}

\item {\bf Claims}
    \item[] Answer: \answerNA{}
    \item[] Justification: Abstract and introduction accurately state contributions supported by experiments.

\item {\bf Limitations}
    \item[] Answer: \answerYes{}
    \item[] Justification: N/A

\item {\bf Theory assumptions and proofs}
    \item[] Answer: \answerNA{}
    \item[] Justification: Primarily empirical; no formal theoretical claims are made.

    \item {\bf Experimental result reproducibility}
    \item[] Answer: \answerYes{}
    \item[] Justification: Full details in Section~4 and Appendices~A--D.

\item {\bf Open access to data and code}
    \item[] Answer: \answerNo{}
    \item[] Justification: Code will be released upon acceptance. All benchmarks are public.

\item {\bf Experimental setting/details}
    \item[] Answer: \answerNA{}
    \item[] Justification: N/A

\item {\bf Experiment statistical significance}
    \item[] Answer: \answerNA{}
    \item[] Justification: N/A

\item {\bf Experiments compute resources}
    \item[] Answer: \answerNA{}
    \item[] Justification: N/A

\item {\bf Code of ethics}
    \item[] Answer: \answerYes{}

\item {\bf Broader impacts}
    \item[] Answer: \answerNA{}
    \item[] Justification: N/A

\item {\bf Safeguards}
    \item[] Answer: \answerNA{}

\item {\bf Licenses for existing assets}
    \item[] Answer: \answerYes{}
    \item[] Justification: All cited; Qwen3 under Apache 2.0.

\item {\bf New assets}
    \item[] Answer: \answerNA{}
    \item[] Justification: No.

\item {\bf Crowdsourcing and research with human subjects}
    \item[] Answer: \answerNA{}

\item {\bf Institutional review board (IRB) approvals or equivalent for research with human subjects}
    \item[] Answer: \answerNA{}

\item {\bf Declaration of LLM usage}
    \item[] Answer: \answerYes{}
    \item[] Justification: Qwen3 usage fully described.

\end{enumerate}

\end{document}